\documentclass[10pt,twocolumn,letterpaper]{article}

\usepackage{iccv}
\usepackage{times}
\usepackage{epsfig}
\usepackage{graphicx}
\usepackage{amsmath}
\usepackage{mathtools,amsfonts,bm}
\usepackage{tabularx}
\usepackage{algorithmicx}
\usepackage[ruled]{algorithm}
\usepackage[noend]{algpseudocode}
\usepackage{float}
\usepackage{graphicx}
\usepackage{gensymb}
\usepackage{amssymb}
\usepackage{mhchem}
\usepackage{lipsum}
\usepackage{float}
\usepackage{xcolor}
\usepackage{caption}

\usepackage{amssymb}
\usepackage{subfigure}
\usepackage{multirow}
\usepackage{booktabs}
\usepackage[pagebackref=true,breaklinks=true,letterpaper=true,colorlinks,bookmarks=false]{hyperref}

\iccvfinalcopy 

\def\iccvPaperID{3} 

\definecolor{purple}{rgb}{0.858, 0.188, 0.478}

\def\mW{{\bm{W}}}

\DeclareMathAlphabet{\mathsfit}{\encodingdefault}{\sfdefault}{m}{sl}
\SetMathAlphabet{\mathsfit}{bold}{\encodingdefault}{\sfdefault}{bx}{n}
\newcommand{\tens}[1]{\bm{\mathsfit{#1}}}
\def\tW{{\tens{W}}}

\newcommand{\etens}[1]{\mathsfit{#1}}

\def\etF{{\etens{F}}}

\def\mW{{\bm{W}}}
\makeatletter
\def\hlinewd#1{%
  \noalign{\ifnum0=`}\fi\hrule \@height #1 \futurelet
   \reserved@a\@xhline}
\DeclareMathOperator{\rank}{rank}

\makeatletter
\newcommand{\multiline}[1]{%
  \begin{tabularx}{\dimexpr\linewidth-\ALG@thistlm}[t]{@{}X@{}}
    #1
  \end{tabularx}
}
\makeatletter
\def\hlinewd#1{%
  \noalign{\ifnum0=`}\fi\hrule \@height #1 \futurelet
  \reserved@a\@xhline}

\ificcvfinal\fi

\begin{document}

\title{CONet: Channel Optimization for Convolutional Neural Networks}

\author{Mahdi S. Hosseini$^{1}$\thanks{Equally major contribution},~Jia Shu Zhang$^{2}$\footnotemark[1],~Zhe Liu$^{2}$\footnotemark[1],~Andre Fu$^{2}$\footnotemark[1],~Jingxuan Su$^{2}$, Mathieu Tuli$^{2}$\\
Sepehr Hosseini$^{2}$\thanks{Equal contribution},~Arsh Kadakia$^{2}$\footnotemark[2], Haoran Wang$^{2}$\footnotemark[2] and Konstantinos N. Plataniotis $^2$\\
$^1$The Department of Electrical and Computer Engineering, University of New Brunswick \\
$^2$The Edward S. Rogers Sr. Department of Electrical \& Computer Engineering, University of Toronto\\
\tt\color{purple}\url{https://github.com/mahdihosseini/CONet}
}

\maketitle
\ificcvfinal\thispagestyle{empty}\fi

\begin{abstract}
   Neural Architecture Search (NAS) has shifted network design from using human intuition to leveraging search algorithms guided by evaluation metrics. We study channel size optimization in convolutional neural networks (CNN) and identify the role it plays in model accuracy and complexity. Current channel size selection methods are generally limited by discrete sample spaces while  suffering from manual iteration and simple heuristics. To solve this, we introduce an efficient dynamic scaling algorithm -- CONet -- that automatically optimizes channel sizes across network layers for a given CNN. Two  metrics -- ``\textit{Rank}'' and ``\textit{Rank Average Slope}'' -- are introduced to identify the information accumulated in training. The algorithm dynamically scales channel sizes up or down over a fixed searching phase. We conduct experiments on CIFAR10/100 and ImageNet datasets and show that CONet can find efficient and accurate architectures searched in ResNet, DARTS, and DARTS+ spaces that outperform their baseline models.
	
This document superceeds previously published paper in ICCV2021-NeurArch workshop. An additional section is included on manual scaling of channel size in CNNs to numerically validate of the metrics used in searching optimum channel configurations in CNNs.
\end{abstract}

\vspace{-0.6cm}
\section{Introduction}\label{sec_introduction}
In the midst of Neural Architecture Search (NAS) algorithms \cite{ZophICLR2017, zoph2018learning, tan2019mnasnet, wang2020neural, liu2019darts, Zela2020Understanding, vahdat2020unas, wan2020fbnetv2, Cai2020Once}, the generated CNN are poorly optimized due to the lack of automated scaling of channel sizes (aka the number of filters in a convolution layer). Given a dataset and a model architecture, the selection of proper channel sizes has a direct impact on the network's success. The over-parameterization of a network can (a) further complicate the training process; and (b) generate heavy models for implementation, while the under-parameterization of a network yields low performance.

\begin{figure}[t!]
\centering
\subfigure[CONet Overview]{\includegraphics[width=1.0\linewidth]{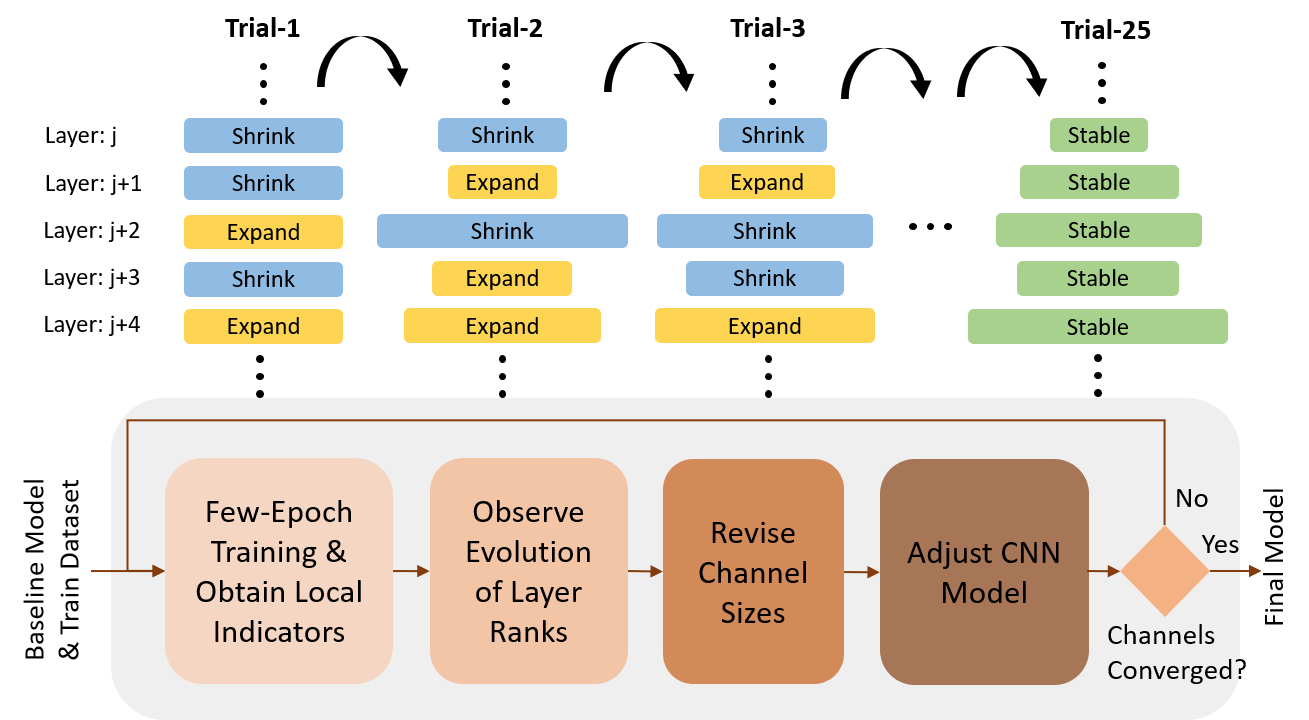}}
\subfigure[DARTS7 Layer 12]{\includegraphics[width=0.2\textwidth]{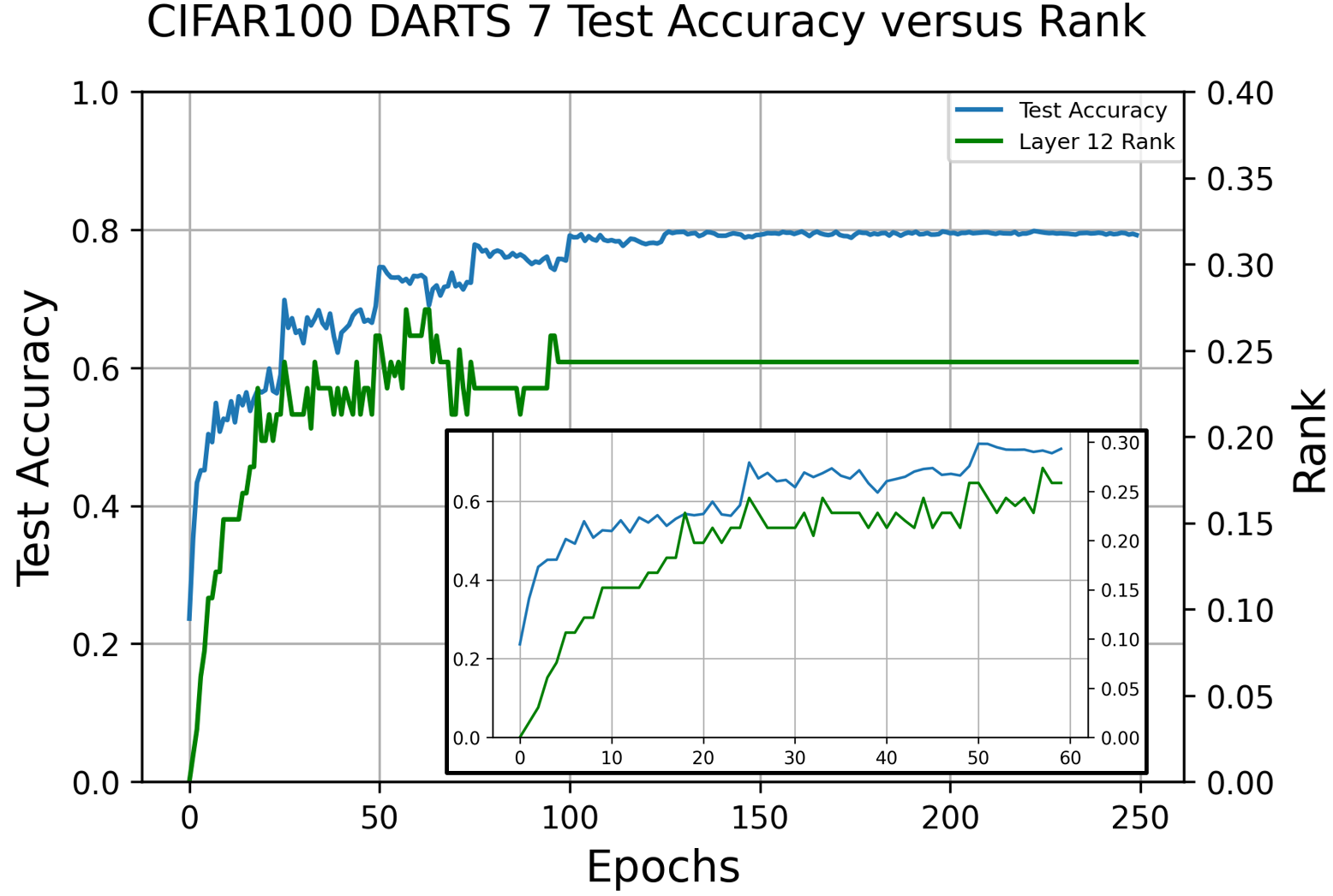}}
    \subfigure[Test Acc vs. Rank]{\includegraphics[width=0.15\textwidth]{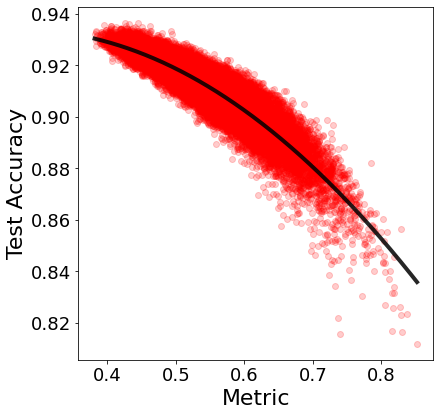}}
\caption{(a) CONet Overview. The algorithm begins with a \textit{small} baseline model and trains a few epochs using the train dataset. Every channel dynamically shrinks/expands based on the rank evolution. This is repeated until all channel sizes have stabilized. (b) We randomly inspect the 12th layer of DARTS7 and over 250 epochs with a zoomed version to 60 epochs in bottom right; (c) Test accuracy vs. Rank on CIFAR10 for the NATS-Benchmark \cite{dong2021nats}. Probing each layer in aggregate allows us to correlate rank with accuracy, allowing optimization of rank as a proxy for test-accuracy.
\vspace{-0.6cm}}
\label{fig:Concept_Figure}
\end{figure}



Few studies show the importance of channel scaling in architecture search. Searching techniques in \cite{tan2020efficientnet, stamoulis2019single, Cai2020Once, wan2020fbnetv2} use a semi-automated pipeline to generate sub-optimal block architectures to fit training data. Their use of heuristics and associated hyper-parameters to search for channel size blocks them from full automation. In this paper we question whether the generated architectures are optimized to obtain a better trade-off between performance, accuracy and model complexity. 

We address the above limitations by introducing a fully automated channel optimization algorithm -- \textit{``CONet''} -- to search for optimal  channel sizes in a network architecture. See Fig. \ref{fig:Concept_Figure}(a) for the overall workflow. The proposed method depends auxiliary indicators that locally probe intermediate layers of cell architectures and measure the utility of learned knowledge during training. Our metrics employ low-rank factorization of tensor weights in CNN and compute; (a) \textit{rank} to determine the well-posedness of the convolution mapping space; (b) \textit{condition} to compute the numerical stability of layer mapping with respect to minor input perturbations; and (c) \textit{rank-slope} to compute the dimension growth of convolution weights. The rank-slope is measured after a few training epochs to evaluate how well a convolution layer is learning. We translate this into a decisive action of shrinking or expanding the channel size for the next evolving architecture. CONet is able to dynamically scale baseline models to achieve higher top-1 test accuracy with minimal parameter count increase, and in some cases, a reduction in parameter count.

The summary of our contributions is listed as follows:

\textbf{1) Metric Development.} We develop new metric tools that can probe intermediate layers of CNN architectures as a strong response function for channel size selection and measure how well the layers are training.

\textbf{2) Independent Channel Allocation.} We design an automated algorithm to find independent connections in CNN and show how individual convolution layers can be adjusted independent of its neighbouring blocks.

\textbf{3) CONet.} We propose channel optimization method: CONet and show how the algorithm dynamically scales channel sizes to convergence.


\textbf{4) Experiments.} Comprehensive experiments are conducted across three architecture spaces (i.e. ResNet \cite{he2015deep}, DARTS \cite{liu2019darts} and DARTS+ \cite{liang2020darts}) and three datasets (CIFAR10, CIFAR100, and ImageNet) and show how the original architectures are dynamically scaled to achieve superior performances compared to their baselines.

\subsection{Related Works}\label{sec:related_works}
\textbf{Hand-Crafted Networks.} Hand-crafted networks are heuristically defined based on expert domain knowledge. Examples include VGG \cite{simonyan2014very}, ResNet \cite{he2015deep}, Inception \cite{szegedy2016rethinking}, ResNeXt \cite{xie2017aggregated}, and MobileNet \cite{sandler2018mobilenetv2}. The rule-of-thumb is to increase the channel size by increasing the network depth layers for better image class representation. 

\textbf{Neural Architecture Search (NAS).} Automatic NAS was first introduced in \cite{ZophICLR2017} to generate efficient (child) networks that achieve the performance of hand-crafted models. NAS is mainly guided by recurrent neural networks (RNN) and trained with Reinforcement Learning (RL) -- known as NAS-RL. Examples are transfer learning method from NASNet \cite{zoph2018learning}, model latency incorporation from MnasNet \cite{tan2019mnasnet}, Q-Learning methods used in \cite{baker2016designing, zhong2018practical}, and child node weight-sharing used in \cite{pmlr-v80-pham18a, YuECCV2020BigNAS, wang2020neural}. All of these methods sample the channel size per layer from a discrete sample set. 

\textbf{Differentiable NAS (DNAS).} A drawback of previous methods is that they search through a discrete space (which is limiting) and tend to take several hundreds to thousands of GPU hours to search. Differentiable Architecture Search (DARTS) \cite{liu2019darts}  was recently introduced to relax this search space by defining auxiliary variables over cell operations and optimizing the architecture design through a bilevel optimization problem. Variants of DARTS have been proposed including PC-DARTS \cite{xu2019pc} to remove redundant searches, Robust-DARTS \cite{Zela2020Understanding} to stabilize the search space, Fair-DARTS \cite{FairDARTSChu2020} to relax the exclusive competition between skip-connections, SNAS \cite{xie2018snas} that formulates a joint distribution parameter within a DARTS cell, UNAS \cite{vahdat2020unas} that unifies DARTS with RL, Progressive-DARTS \cite{chen2019progressive} that progressively searches the network depth, and SGAS \cite{li2020sgas} that divides the search procedure and applies pruning. These methods start with a heuristic selection of fixed channel sizes and scale up after each reduction cell. None-DARTS DNAS method also exist, including FBNetV2 \cite{wan2020fbnetv2} and Single-Path NAS \cite{stamoulis2019single}, however they suffer from similar drawbacks as DARTS, although Single-Path NAS does improve in search time. Overall, the major drawback of DNAS is the requirement of expert domain knowledge for tuning hyper-parameters for training.

\textbf{Other methods.} Pruning methods exist, including TAS \cite{dong2019network}, DF \cite{li2019partial}, OFA \cite{Cai2020Once}, AtomNAS \cite{Mei2020AtomNAS}, and ASAP \cite{noy2020asap}. One major drawback of the above pruning methods is that the initial large model requires extensive computational resources for training. Network search based on evolutionary algorithms also exist, include binary encoding used in GeNet \cite{xie2017genetic}, mutation algorithms for large-space searching used in Large Scale Evolution \cite{real2017large}, mutation algorithms applied to NASNet used in AmoebaNet-A \cite{real2019regularized}, and RL applied to evolutionary algorithms used in FPNAS \cite{cui2019fast}. In each of these methods, channel sizes follow manual selection heuristics.

\section{Local Indicators for CNN} \label{sec:rank_evaluation}
Central to our approach are three metrics developed to probe intermediate layers of CNN and evaluate the well-posedness of the trained convolutional weights.

\subsection{Low-Rank Factorization}
Consider a four-way array of convolution weight $\tW \in\mathbb{R}^{N_{1} \times N_{2} \times N_{3} \times N_{4}}$, where $N_{1}$ and $N_{2}$ are the kernel height and width, and $N_{3}$ and $N_{4}$ are the input and output channel sizes, respectively. In convolution, the input features ${\etF}^{I}\in\mathbb{R}^{H \times W \times N_{3}}$ are mapped into an output feature map $\etF^{O}_{:,:,i_4} = \sum_{i_3=1}^{N_3}\etF^{I}_{:,:,i_3}\ast\tW_{:,:,i_3,i_4}$, where ${\etF}^{O}\in\mathbb{R}^{H \times W \times N_{4}}$. The convolution mapping acts as an encoder, and therefore we can prob the encoder to understand the structure of weights for convolution mapping. This is useful to understand the evolution of its structure over iterative training. To achieve this, we unfold the tensor into a two-dimensional matrix along a given dimension $d$

\centerline{$\tW{\texttt{[Tensor-4D]}} \xrightarrow{\text{unfold}} {W_d}{\texttt{[Matrix-2D]}}$.}
For instance, the unfolded matrix along the output dimension i.e. $d=4$ is given by $W_4\in\mathbb{R}^{N_{4} \times N_{1}N_{2}N_{3}}$. The weights in CNN are initialized (e.g. random Gaussian) and trained over some epochs, generating meaningful structures for encoding. However, the random perturbation inhibits proper analysis of the weights during training. We decouple the randomness to better understand the trained structure. Similar to \cite{lebedev2015speeding}, we obtain the low-rank structure

\begin{figure}[htp]
\begin{center}
\includegraphics[width=0.8\linewidth]{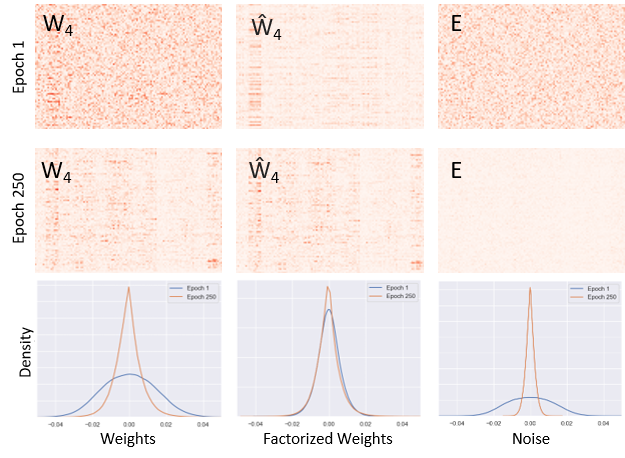}
\end{center}
\caption{Low-rank factorization on a ResNet layer during first and last epoch training.}
\label{fig:low_rank_histogram}
\end{figure}

\centerline{$W_d \xrightarrow{\text{factorize}} \widehat{W}_d + E$,} where, $\widehat{W}_d$ is the low-rank matrix structure containing limited non-zero singular values i.e. $\widehat{W}_d = U \Sigma V^T$, where $\Sigma=\text{diag}\{\sigma_1, \sigma_2, \hdots \sigma_{N^{\prime}_d}\}$ and $N^{\prime}_d = \rank{\widehat{W}_d}$. Here $N^{\prime}_d<N_d$ due to the low-rank property. We then employ Variational Baysian Matrix Factorization (VBMF) method from \cite{nakajima2013global} which factorizes a matrix as a re-weighted SVD structure. The method is computationally fast and can be easily applied to several layers of arbitrary size. Figure \ref{fig:low_rank_histogram} demonstrates the low-rank factorization on a convolution layer of ResNet34 during two different training phases i.e. first epoch and last epoch. Notice how the low-rank structure (aka useful information) is evolved while the noise perturbation is faded at the end; leading to a stabilized layer.

\textbf{Low-Rank Measure.} We define the first probing metric, the \textit{rank}, by computing the relative ratio of the number of non-zero low-rank singular values to the given channel size
\begin{equation}
    R(\widehat{W}_d) = N^{\prime}_{d} / N_{d}.
    \label{eq:rank}
\end{equation}
The rank in Eq. \ref{eq:rank} is normalized such that $R(\widehat{W}_d)\in[0,1]$ and it measures the relative ratio of the channel capacity for convolutional encoding. Note that the unfolding can be done in either input or output dimensions i.e. $d\in \{3, 4\}$ and therefore each convolution layer yields separate input/output rank measurements i.e. $R(\widehat{W}_3)$ and $R(\widehat{W}_4)$.

\textbf{Low-Rank Condition.} We adopt a second probing metric, the \textit{low-rank condition}, to compute the relative ratio of maximum to minimum low-rank singular values
\begin{equation}
\kappa({\widehat{\mW}_d})={\sigma_{1}({\widehat{\mW}}_{d})}/{\sigma_{N^{\prime}_{d}}({\widehat{\mW}}_{d})}.
\label{eq:mapping_condition}
\end{equation}
The notion of matrix condition in Eq. \ref{eq:mapping_condition} is adopted from matrix analysis \cite{horn2012matrix} and measures the sensitivity of convolution mapping to minor input perturbations. If the condition is high for a particular layer, then the input perturbations will be propagated to the proceeding layers leading to a poor auto-encoder.

\subsection{Low-Rank Evolution}\label{sec:rank_evolution}
\begin{figure}[htp]
\begin{center}
\includegraphics[width=1\linewidth]{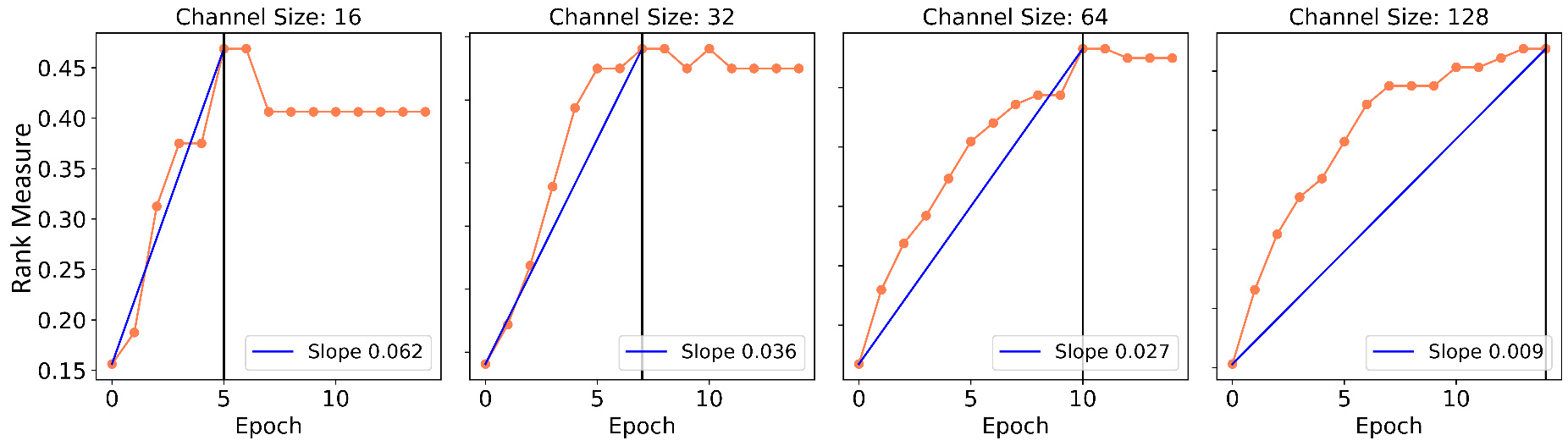}
\end{center}
\caption{Evolution of rank slopes on different initial channel sizes: increasing the channel size, decreases the slope.}
\label{fig:rank_different_conv_size}
\end{figure}

While both Rank (Eq. \ref{eq:rank}) and Condition (Eq. \ref{eq:mapping_condition}) are useful metrics to observe a layer's learned knowledge compared to other layers, they are less sensitive to the selection of channel size. We observe the evolution of the metrics over training, yielding meaningful insights into how channel size selection impacts the convergence of the metric. Shown in Fig. \ref{fig:rank_different_conv_size}, we train the network and observe the rank averages for different channel sizes of the same layer index (i.e. we vary channel size). Notice how the rank slope is inversely related to the increase in channel size.

We define the \textit{rank-slope} by computing the relative ratio of the rank gain to the number of epochs taken to plateau
\begin{equation}
    S = \frac{\overline{R}(\widehat{W}_d^{t_2}) - \overline{R}(\widehat{W}_d^{t_1})}
    {t_2 - t_1},
    \label{eq:rank_slope}
\end{equation}
where, $\overline{R}=[R_{3}+R_{4}]/2$ is the average rank, $t_1$ the starting epoch, and $t_2$ is the rank plateau epoch. The slope, $S$ in Eq. \ref{eq:rank_slope} encodes the complexity of the channel size selection: (a) how much rank (knowledge) is gained within a layer; and (b) the amount of time to reach the maximum rank capacity --indicating a target for channel size selection.

\begin{figure*}[htp]
\begin{center}
\includegraphics[width=0.35\linewidth]{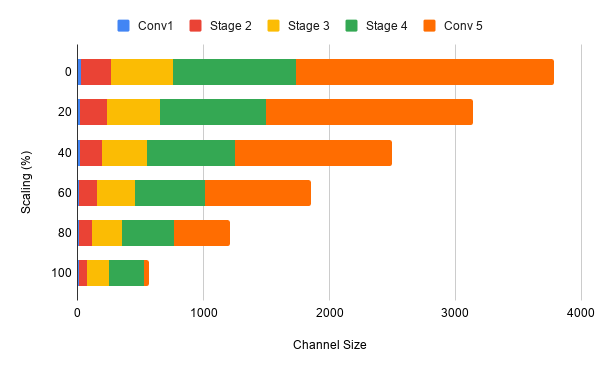}
\includegraphics[width=0.35\linewidth]{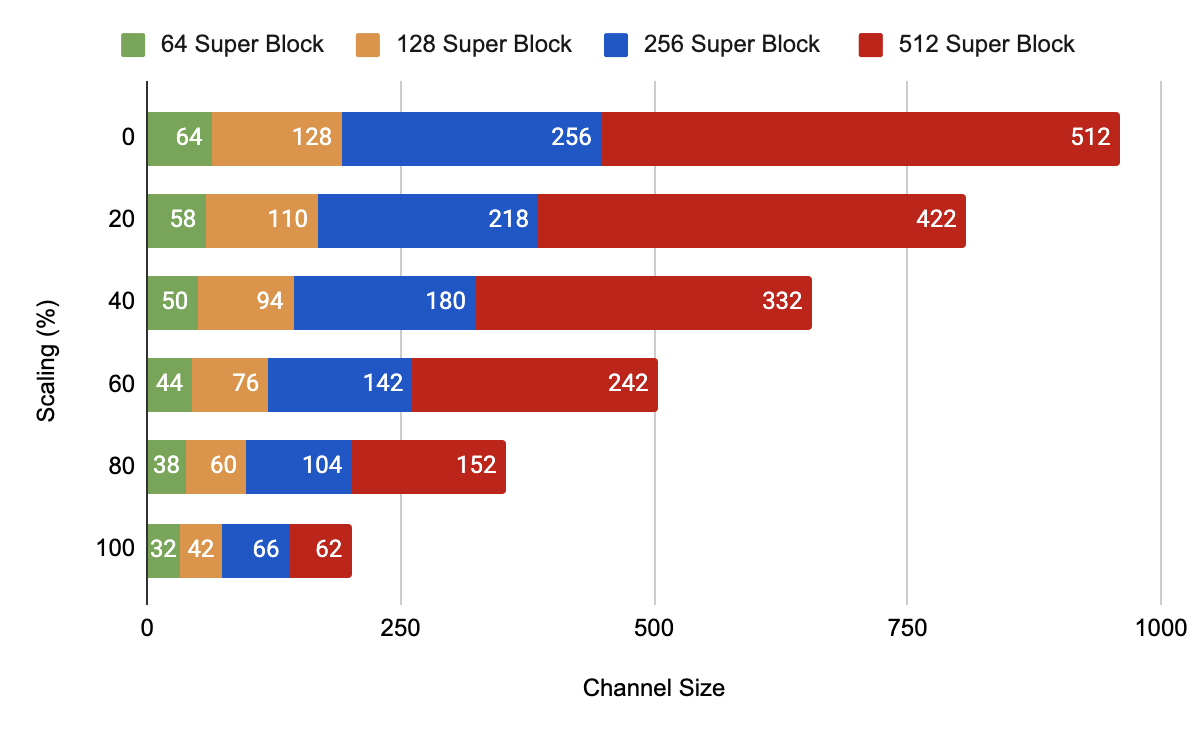}\\
\includegraphics[width=0.27\linewidth]{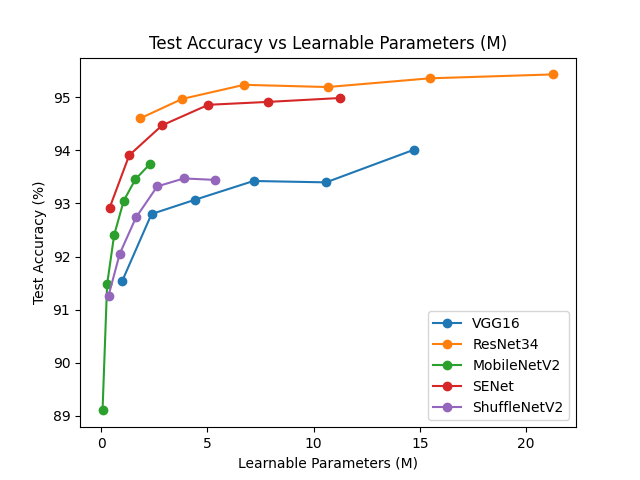}
\includegraphics[width=0.27\linewidth]{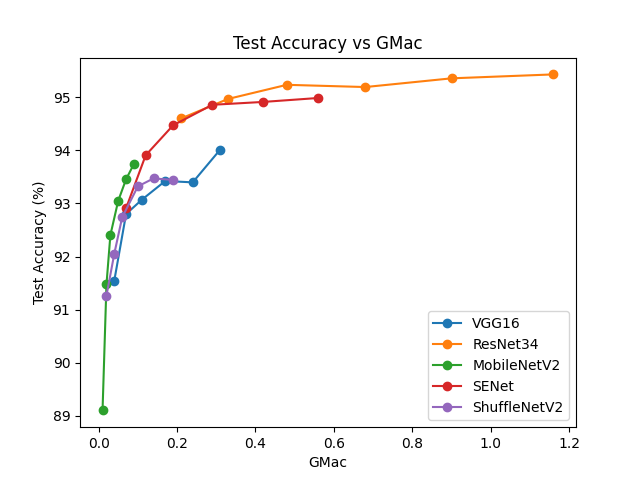}
\includegraphics[width=0.27\linewidth]{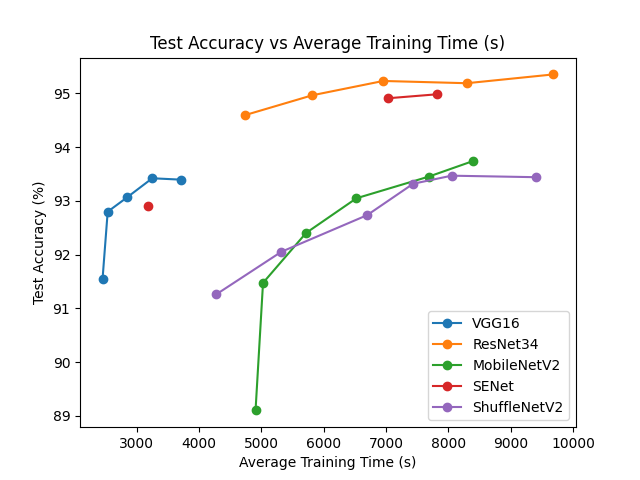}
\end{center}\vspace*{-0.6cm}
\caption{\textbf{Top-left:} visualization of the ShuffleNetV2 \cite{ShuffleNetv2} network architecture with respect to the scaling of channel sizes. Each colour represents the channel size  from a specific conv block in \cite{ShuffleNetv2}. Note that the conv1 layers' channel size is too small to see in higher scalings. As shown above, each channel size decreases significantly as the scaling percentage across the vertical axis increases. The output rank of each individual convolutions are averaged with respect to each stage to obtain the final scaling factor used to produce the respective channel size. \textbf{Top-right:} Visualization of the SENet \cite{SENet} architecture and its respective channel sizes per layer for various degrees of scaling. Each colour represents a convolutional layer with a corresponding channel size, which decreases as channel sizes are gradually reduced with higher degrees of scaling. \textbf{Bottom:} Test accuracies of various networks compared with respect to important network details such as training time, parameter count, and MACs trained on CIFAR10. Each bullet along each respective line corresponds to a specific scaled version of that network ranging from baseline (no scaling) to fully scaled (100\% scaling as informed by output rank metrics.)}
\label{fig:cifar10_acc_graphs}
\end{figure*}
\section{Static Channel Scaling}
In this section we study a simple approach for channel scaling of ConvNets. The first stage in this process is to train the networks without any scaling as a baseline reference for all other scaled results. Next, the maximum output rank for each convolutional layer was extracted from these baseline results for each network. As discussed in previously, the output rank provides a metric for how much knowledge gain is being used, and hence serves as a good guideline for determining what percentage of the channel size is actually saturated. Therefore, the product of the output rank (a ratio) and the baseline channel size for a specific convolutional layer will result in the effective channel size for maximum scaling (referred to as 100\% scaling). To derive as much insight as possible with regards to convolution sizes and the network performance, the networks were gradually scaled down in increments of 20\% (calculated as a factor of the maximum scaling) until the networks were fully scaled. Furthermore, incremented scaling allows us to explore the optimal scaling for various networks. The general formula used can be stated as the following:
\begin{equation}
    \label{eqn:channel_scaling}
    C_{f} = C_{i} \times (O + (1 - S) \times (1 - O)) 
\end{equation}
In Equation \ref{eqn:channel_scaling}, $C_{f}$ is defined as the final, scaled, output channel size, while $C_{i}$ is the initial output size from the baseline network. $O$ refers to the maximum output rank and $S$ is the scaling factor defined in increments of 20\% ranging from 20\% to 100\% scaling. 

It is worth noting that the channel size for a single layer is not necessarily simply based on the maximum output rank for that given layer depending on the architecture. When networks are scaled, maintaining the integrity and architecture of the original network is a priority to reduce unintended effects. Therefore, in all scaled networks, the number of layers and shortcuts remains the same. To allow for this similarity of configuration, blocks that have multiple layers with the same output channel sizes is continued in scaling networks. In this scenario, an average is taken of the output ranks in a set of blocks that all have the same output size. This average is utilised as the final output rank for scaling purposes. For example, in the first ResNet34 "superblock" that contains all layers (excluding shortcut layers) with an output channel size of 64, an average is taken for the output rank of those layers and then applied to Equation \ref{eqn:channel_scaling} to generate a final size.

\subsection{Preliminary Observation}
We performed our scaling experiments on the most popular networks including VGG16, ResNet34 and SENet for maximum utility due to their popularity yet can benefit from the computational reductions as a result of scaling. In addition, the same methodologies were applied to compact mobile networks such as MobileNetV2 and ShuffleNetV2 to explore the effects of scaling on networks that are specifically designed to be low-cost.

We evaluated our scaled networks on the CIFAR10 dataset with 32x32 images and 10 classes. This allowed us to train and obtain results in the most time efficient manner to validate the results of our scaling before moving on to the automated scaling algorithm in next sections. Using the metrics that we collected, we were able to visually assess the effects of the scaling with respect to important model properties and deduce important observations. Through Figure \ref{fig:cifar10_acc_graphs}, it is clear that the ResNet architecture handled scaling the best as various model properties such as GMACs, parameter count, and average training time all showed steady decreases while minimally hindering performance. Furthermore, the fully scaled version of ResNet34 showed superior performance against all other networks while retaining a parameter count equivalent to that of baseline MobileNetV2. As a result, it is clear that ResNet34 is the best network to use on CIFAR10 for optimal performance given any computational budget as all scaled versions retained above 94\% test accuracy. 
\section{Network Connectors}\label{sec:network_connection}
While we can measure the well-posedness of each layer, we are often unable to individually alter channel sizes. This is because changing the channel size of a layer affects the depth of its feature map, which impacts subsequent layers. E.g given a pair of conv-layers in series, the output channel size of the first layer must match the input channel size of the second layer. We  represent a network as a directed acyclic graph (DAG) to identify these constraints. The nodes are layer \textit{depth} (aka \textit{channels}). The edges are operators on the feature maps between the tail \& head.

\subsection{Connection Types}
\begin{figure}[htp]
    \centering
    \includegraphics[width=0.4\textwidth]{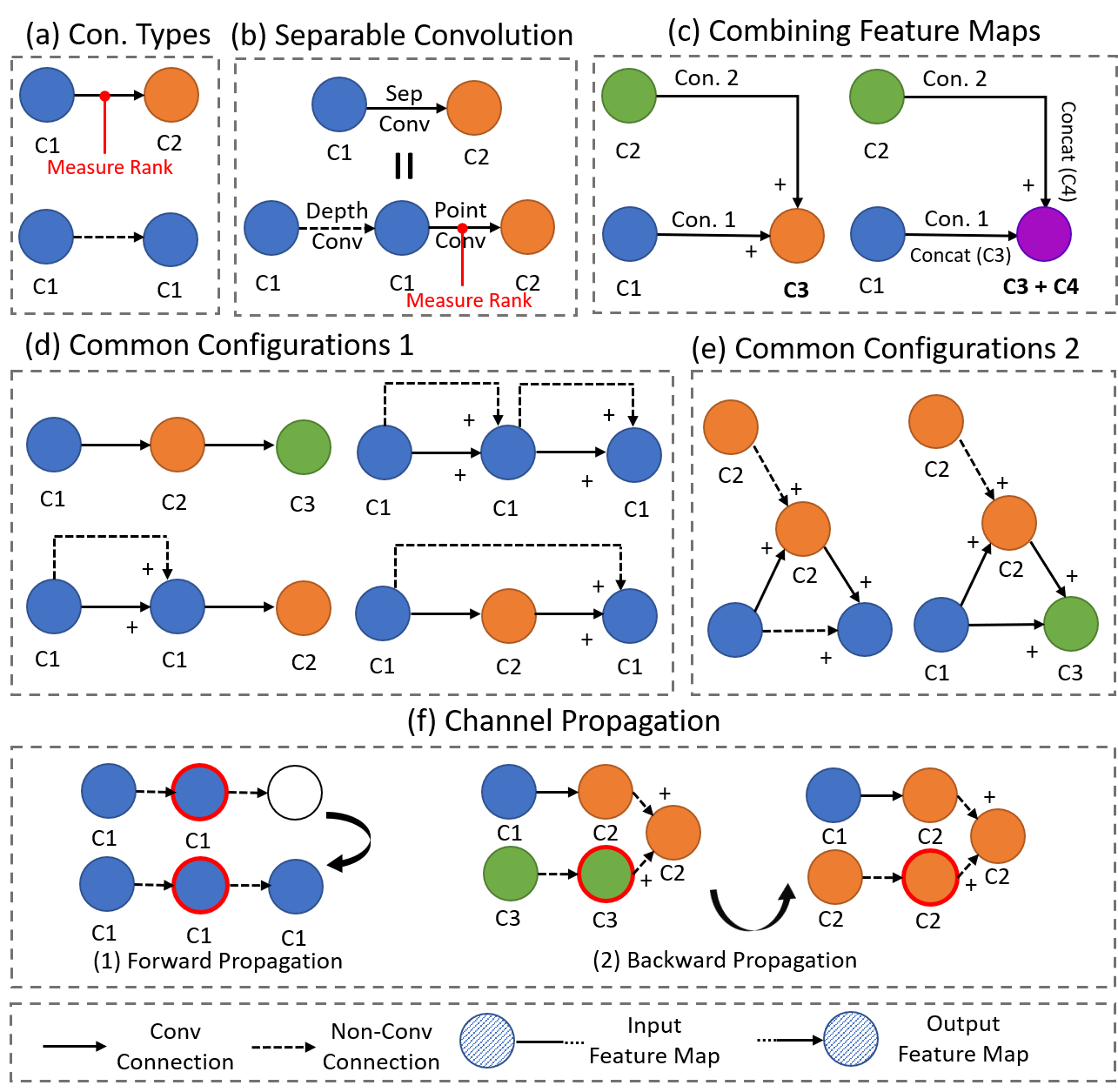}
    \caption{Overview of local connectors in a typical CNN architectures to identify independent channels for probing: (a) Elementary connections; (b) Rank measure on separable convolution; (c) Summation and Concatenation of feature maps; (d) Common configurations within residual networks; (e) Common configuration within inception/cell-based networks; (f) Propagation of Channel Constraints.}
    \label{fig:connection_types}
\end{figure}

In Fig. \ref{fig:connection_types} (a), we categorize the edges into two types: \textit{conv} and \textit{non-conv}. \textit{Conv} edges represents convolution operations and \textit{non-conv} edges represents other operations e.g skip-connection and max-pooling. There are 2 distinctions between the these edge types. (1), only conv edges can connect nodes with different depths. Non-conv edges must have the same depth for the input and output nodes. (2), we measure  rank from conv edges to adjust the channel sizes of a network. 

We further define an edge is \textit{inbound} to a node if it is connected at the head of the edge. The edge is \textit{outbound} to a node if it is connected at the tail of the edge. In most cases, the depth of the node determines the input channel size for all outbound edges and the output channel size for all inbound edges. The exception is the case for concatenated nodes, which will be described later.

\textbf{Separable Convolutions.} A separable convolution is comprised of a depthwise layer and pointwise layer in series. The depthwise layer contains a filter for each channel of the input feature map. Each filter has a depth of 1 and convolves only with a single channel. The output from each filter is concatenated and passed to the pointwise layer. The pointwise layer contains 1x1 kernels that combine the concatenated output along the depth dimension.

As show in in Fig. \ref{fig:connection_types} (b), we treat the depthwise layer as a non-conv edge. This is because the input and output channel sizes for a depthwise layer must be the same since each filter only process a single channel from the input to produce a single channel in the output. Furthermore, we found that the rank of depthwise layers behaves unpredictably because each filter only has a depth of 1. We  use the ranks of the pointwise layer to adjust the entire separable convolution.

\textbf{Summation and Concatenation.} When multiple edges are inbound to the same node, the output from each edge must be combined. The two ways of combining feature maps are \textit{summation} and \textit{concatenation}. Fig. \ref{fig:connection_types} (c) demonstrates the effect on channel sizes for each method. For summation, the output channel size of each edge must be the same to avoid dimension mismatch. For concatenation, the output channel size of each edge can be different. The final depth of the concatenated node will be the total output channel size across all inbound edges.

\textbf{Example 1.} The configurations shown in Fig. \ref{fig:connection_types} (d) are commonly found in network architectures that uses repeating skip-connections such as ResNet \cite{he2015deep} and DenseNets \cite{huang2017densely}. The configurations show that skip-connections can propagate channel size constraints across the network. 

\textbf{Example 2.} The configurations shown in Fig. \ref{fig:connection_types} (e) are commonly found in inception \cite{he2015deep} and cell-based \cite{real2019regularized} \cite{liu2018darts} network architectures. One simple way of identifying if the depth of a node is independent is by checking if all inbound edges are conv connections. 

\begin{algorithm}[H]
    \caption{Unique Channel Size Assignment}
    \label{alg:channelalg}
\textbf{Input: } DAG Network\\
\textbf{Output: } Unique channels
\footnotesize
\begin{algorithmic}[1]
    \State Initialize $node$ as unassigned $\forall$ nodes \hfill 
    \State Store nodes in $queue$ in partial order \hfill  // Sec. \ref{sec:channel_constraints}
    \For{$node$ \textbf{in} $queue$}
        \If{$node$ $=$ unassigned}
            \If{$summation$}
                \State channel $\leftarrow$ unique size
            \ElsIf {$concatenation$}
                \State channel $\leftarrow$ sum of $inbound$ $edges$
            \EndIf
        \EndIf
        \For{$edge$ \textbf{in} $outbound$ $edges$}
        \If{$edge \neq conv$}
            \If{$connected$ $node$ $=$ unassigned}
                \State \multiline{%
                Forward Propagate channel //Fig. \ref{fig:connection_types} (f.1)} 
            \Else 
                \State \multiline{%
                Back Propagate channel \hfill //Fig. \ref{fig:connection_types} (f.2)}
            \EndIf
        \EndIf
        \EndFor
    \EndFor
\end{algorithmic}
\end{algorithm}

\subsection{Channel Size Constraints}\label{sec:channel_constraints}
Using the DAG network representation outlined in Sec. \ref{sec:network_connection}, we can systematically identify channel size constraints for any CNN. We revisit each node in the network sequentially and propagate any channel size constraints caused by non-conv connections throughout the network. If a node has no inbound non-conv connections, then that means it has no channel size constraints with respect to the previous nodes. The overall methodological flow is presented in Alg. \ref{alg:channelalg}. The steps of the flow is also described below.

\textbf{Initializing the Network. } We initialize all network channels as \textit{unassigned}. We then queue the nodes of the network in partial order. For any node in the queue, its outbound edges can only connect to nodes later in the queue.

\textbf{Assigning Channel Size for a Node:} If the channel of a node is unassigned when we pull the node from the queue, we can assign new unique channel sizes to the network. Having an unassigned node implies that all inbound edges are conv connections. If the inbound edges are summed at the node, then we can set a new output channel size shared by all inbound edges. If the inbound edges are concatenated at the node, then each inbound edge can have an independent output channel size.

\textbf{Resolving Channel Size Conflicts: } We propagate the channel size of the current node to subsequent nodes connected by outbound non-conv edges. If we find that the connected node is already assigned, then we must back-propagate that channel assignment through all non-conv edges. See Fig. \ref{fig:connection_types} (f.2) for an illustration of this scenario.

\section{CONet: A Dynamic Optimizing Algorithm}\label{sec:co_net}

\subsection{Important Variables}
\begin{table}[H]
\begin{center}
\bgroup
\def\arraystretch{1.1}
\vspace{-0.1cm}
\footnotesize
{\centering\hfill
\begin{tabular}{p{0.5cm}p{7cm}}
$R$ & List of input and output ranks for each layer.\\
$\kappa$ & List of input and output Mapping condition.\\
$\overline{R}$ & Rank averages.\\
$\overline{\kappa}$ & Mapping condition averages. \\
$S$ & Rank average slopes. \\
$\delta$ & \textit{Rank average slopes threshold.}\\
$\mu$ & \textit{Mapping condition threshold.}\\
$C_{\ell}^{o}$ & Old channel sizes, per layer ($\ell$).\\
$C_{\ell}^{n}$ & New channel sizes, per layer ($\ell$).\\
$\omega_{\ell}$ & Last operation list, per layer ($\ell$).\\
$\phi $ & Scale factors of channel sizes.\\
$\gamma$ & \textit{Stopping condition for scaling factor.} \\
\end{tabular}\hfill}
\egroup
\\
\vspace{0.2cm}
    \caption{Important Variables referenced in Alg. \ref{alg:coalg}. A full list of all variables and their annotations can be found in the supplementary materials. Italicized variables are the 3 key hyper-parameters outlining this algorithm. }
    \label{tab:glossary}
\end{center}
\end{table}
\normalsize
\vspace{-5mm}
We label some key variables used for Alg. \ref{alg:coalg} in Table \ref{tab:glossary} and highlight the following: (1) $\delta$ - threshold rank average slopes, a new hyper-parameter we are introducing, allowing the network to autonomously search with $1$ key hyper-parameter, Sec. \ref{sec:shrink_expand_channel}. (2) $S$ - rank average slope (Eq. \ref{eq:rank_slope}) - allowing probing of how well $R$ is increasing (i.e. how well the model is learning). Finally, (3) $\phi$ - Channel size factor scale, allows the network to continuously alter the channel sizes until convergence.

\subsection{Channel Shrinkage/Expansion} \label{sec:shrink_expand_channel}
\textbf{Calculation of rank average slopes:} We calculate the rank average slopes $S(\ell)$ (Eq. \ref{eq:rank_slope}) for every convolutional layer, $\ell$, as described in Section \ref{sec:rank_evolution}. This slope represents the rate at which $R$ grows as the layer learns. 

\textbf{Criteria for shrinking or expanding channel sizes:} Using rank average slopes $S(\ell)$ we define a target threshold slope, $\delta$. If $S > \delta$ we expand the channel size as it indicates $R$ is growing quickly and as such would benefit from a larger channel size, we then assign the action $\omega_{\ell} = +1$ for expansion. If $S < \delta$ we shrink the channel size as it indicates $R$ is not increasing rapidly so additional channel sizes mean an increase in parameters that do not contribute to model performance. We then assign the action $\omega_{\ell} = -1$ for shrinking. If $S = \delta$ or a no-operation decision was taken we assign $\omega_{\ell} = 0$.

\textbf{Calculation of new channel size:} After a decision to shrink or expand the channel size has been determined, we change the scale the channel size by a factor $\phi_\ell$. Initially we set this scale $\phi_\ell\leftarrow\lfloor 20\% \rfloor$. The new channel size $C_{\ell}^n$ are then computed as a factor scale of the old channel size $C_{\ell}^o$.
\begin{equation}
\label{eqn:channel_size_equation}
   C^n_{\ell} = C^o_{\ell} (1+\omega_{\ell} \cdot \phi_{\ell})
\end{equation}
At the end of every trial a new channel size for every layer is calculated and implemented. If the channel size begins to oscillate between $C_{\ell}^n$ and $C_{\ell}^o$ it indicates the optimal channel size is between the two channel sizes and so we half the factor scale continuously until channel size converges and a no-operation decision is taken.

\subsection{CONet Algorithm}\label{sec:DySNet_Algorithm}
\textbf{Intuition \& Overview:} CONet algorithm\footnote{A PyTorch package is submitted in the supplementary materials.} shown in Alg. \ref{alg:coalg}, begins with a small network using standard channel sizes (e.g $32$) for all layers in the network. We begin a trial, training on the training-dataset, for a predefined number of epochs while recording the average rank per epoch, see Fig. \ref{fig:rank_different_conv_size}. After the trial ends, we stop training, and analyze the rank average slope, $S$ comparing the slope for that layer to our hyper-parameter $\delta$. As per Sec. \ref{sec:shrink_expand_channel} we then take a shrink or expand decision and calculate a new channel size per Eq. \ref{eqn:channel_size_equation}. We continuously execute trials until a predefined number of trials is reached where the channel sizes have converged. By only analyzing $S$ growth, our method never sees the test dataset, thereby optimizing solely on information gained from training data.

\begin{algorithm}[H]
    \caption{Channel Size Optimizaion}
    \label{alg:coalg}
\textbf{Input: } Network, hyperparameters $\delta, \gamma, \mu, \phi $\\
\textbf{Output: } Optimized Network
\footnotesize
\begin{algorithmic}[1]
    \For{trial \textbf{in} trials}
        \State Train current model for a few epochs
        \State Compute average rank ($\overline{R}$) per layer \hfill // Eq. \ref{eq:rank}
        
        \State Compute average MC ($\overline{\kappa}$) per layer \hfill // Eq. \ref{eq:mapping_condition}
        \State From $\overline{R} $ get rank-slope ($S$) per layer \hfill // Eq. \ref{eq:rank_slope}
        \For{layer \textbf{in} network}
           \State Set action to Expand 
            \If{$S[$layer$] < \delta $ or $\overline{\kappa}[$layer$] > \mu$}
                \State Set action to Shrink
            \EndIf
            \If{Last action $\neq$ current action}
                \If{$\phi < \gamma$}
                    \State Stop, channel sizes have converged
                \EndIf
                \State $\phi \leftarrow \phi$ /2    
            \EndIf
            \State Calculate New Channels  \hfill // Eq. \ref{eqn:channel_size_equation}
        \EndFor
        \State Instantiate model with new channels
    \EndFor
\end{algorithmic}
\end{algorithm}

\vspace{-0.3cm}
\begin{figure}[htp]
    \centering
    \includegraphics[width=0.8\linewidth]{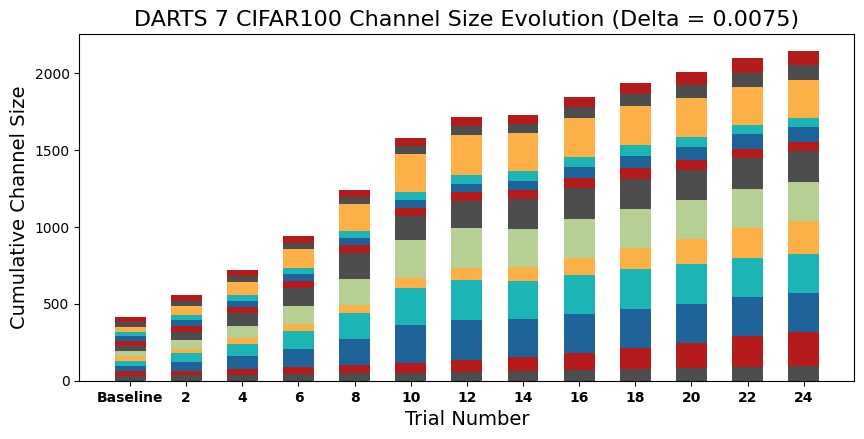}
    \caption{Channel size evolution example of DARTS7 using CONet algorithm on CIFAR100. Channel sizes from only the last cell are shown for better visualization. Specific channel sizes may expand or shrink but the network as a whole stabilizes by the last trial.}
    \label{fig:channel_evolution}
\end{figure}


\textbf{Delta threshold tuning:} We can vary the model parameter count and accuracy trade-off through the $\delta_S$ hyper-parameter. Once an initial $\delta$ is chosen, test values around it and observe the performance change. The $\delta$ values we tested are listed under Sec. \ref{sec:results}.


\textbf{Epochs \& Trials:}  The number of epochs per trial is given from number of epochs to rank convergence. For DARTS and DARTS+, we recommend $20$ epochs per trial is optimal for convergence. The number of trials is given from number of trials to channel size convergence. For DARTS and DARTS+ $25$ trials is optimal for convergence. Results using these hyper-parameters have been described in Sec. \ref{sec:results}. 

\textbf{Algorithm channel size updates:} The algorithm collects the metrics from the first trial, then computes $S$ for each layer (Sec. \ref{sec:shrink_expand_channel}). The decision to shrink or expand the channels are given in Sec.\ref{sec:shrink_expand_channel}. The ability to expand and shrink the channel sizes dynamically is a key advantage over other methods that only scale up \cite{tan_effnet_2019} with no consideration of the  accuracy-parameter trade-off. The algorithm then returns the new channel sizes, $C_{\ell}^n$.


\section{Experiments}
\textbf{Experiment Setup.} We complete experiments on ResNet, DARTS$7$ ($7$ cells), DARTS$14$ (14 cells), DARTS+$7$ and DARTS+$14$. Each model is applied to CIFAR10, CIFAR100 \& ImageNet \cite{krizhevsky2009learning, deng2009imagenet}. All hyper-parameter choices and further implementation details can be found in Supplementary material. 

\textbf{Channel Searching.} 
For the chosen search spaces, we perform a channel search for each model and respective (training set) datasets, with the following thresholds: $\delta$ = $\{\delta_0, \delta_1,\delta_2, \delta_3\}$, 
and for DARTS$14$ and DARTS+$14$, $\delta$ = $\{\delta_0, \delta_1,\delta_2, \delta_3, \delta_4\}$. Each channel search runs for $25$ trials each with $20$ epochs per trial.  


\textbf{Model evaluation.} We evaluated both, the CONet optimized model and unaltered baseline model on CIFAR10/100. For the ImageNet dataset, we use the DARTS$14$ model searched based on CIFAR100 with $\delta =\{\delta_2, \delta_3\}$ and the baseline model of DARTS$14$ from the DARTS paper\cite{liu2018darts}. The baseline models were run on 2 Nvidia RTX 2080Tis while the other SOTA results were taken from their respective papers. 

\subsection{Results}\label{sec:results}

\begin{figure*}[htp!]
\vspace*{-0.6cm}
\centering
    \subfigure[CIFAR10]{\includegraphics[height=0.25\textwidth]{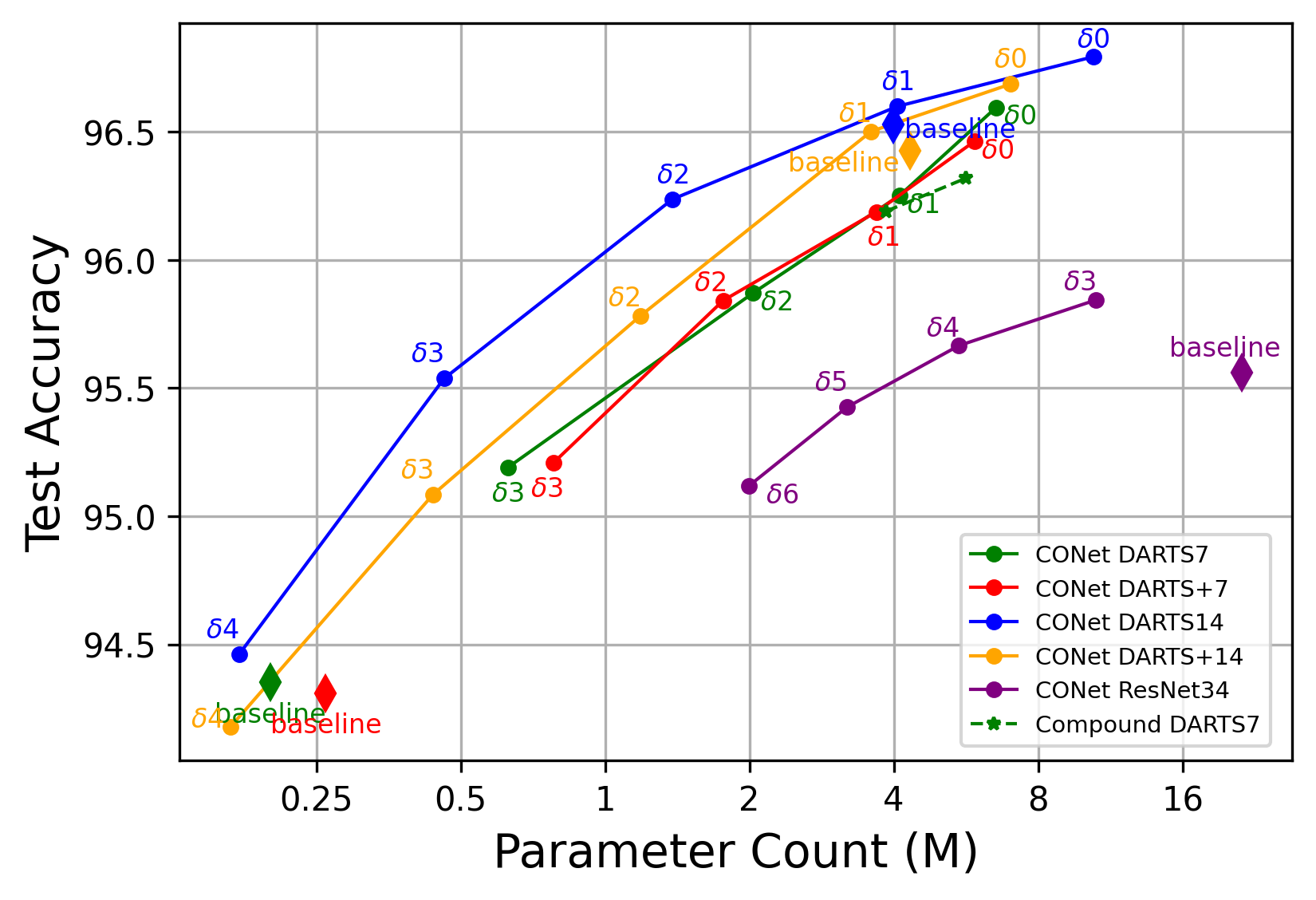}}
    \subfigure[CIFAR100]{\includegraphics[height=0.25\textwidth]{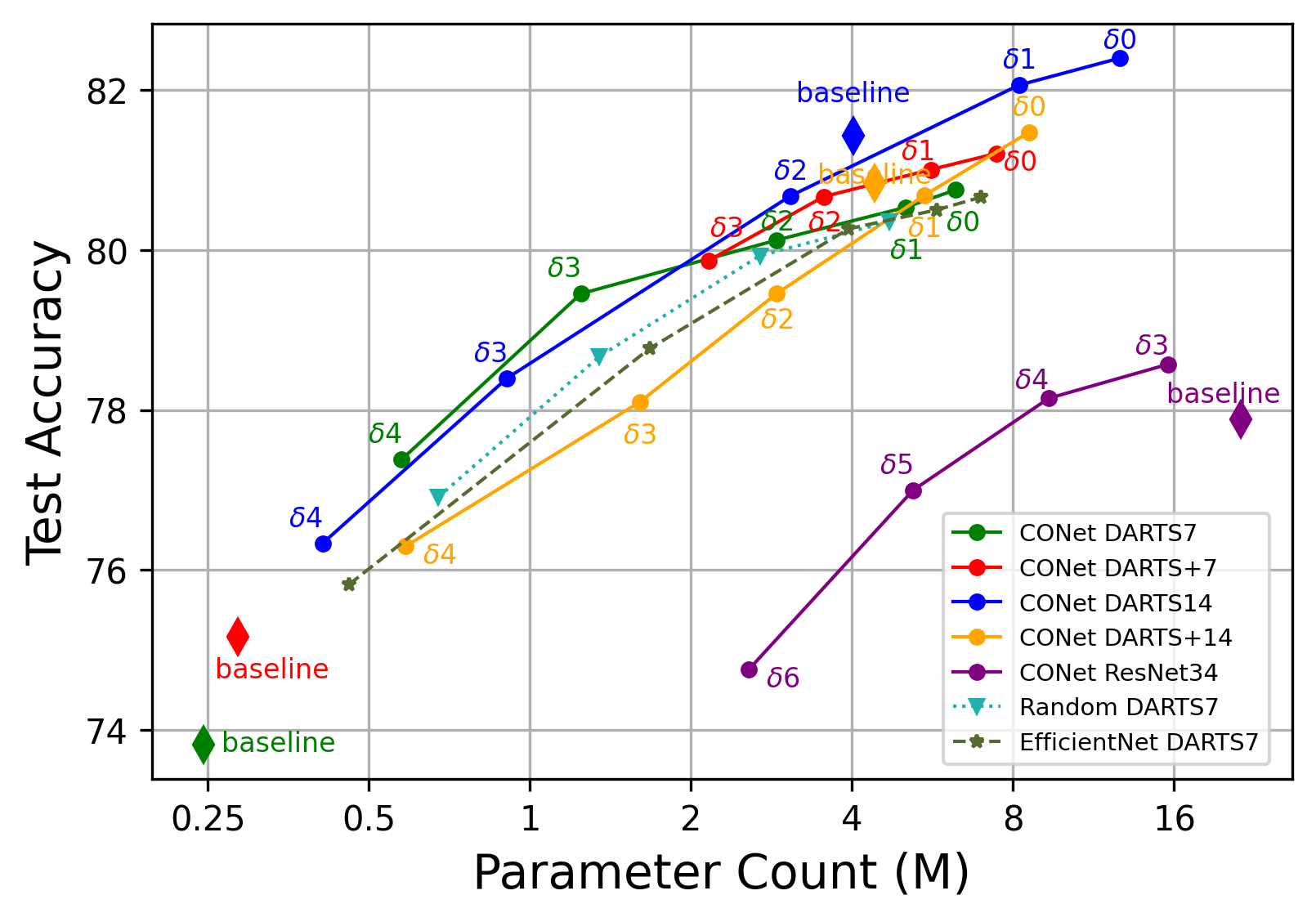}}
\caption{Testing accuracy vs. parameter count for CONet experiment applied on (a) CIFAR10 and (b) CIFAR100. Note that the baselines results are plotted as a diamond scatter plot, and sequential deltas for the CONet are plotted in line graphs with the annotated delta thresholds. Also, the models produced by compound scaling are plotted in dashed line graphs.}
\label{fig:size_vs_acc}
\end{figure*}


We implement CONet within the DARTS, DARTS+ and ResNet search spaces. We compare our method to the Efficient Net scaling algorithm \cite{tan_effnet_2019} and show that the ability to dynamically scale up and down the channel sizes has merit over simple scaling factors as it doesn't consider knowledge growth, redundant channel sizes \& the parameter-accuracy trade-off. There are two phases per experiment, (1) the channel searching phase, searched on training data, and (2) the dataset evaluation phase, evaluated on test data.  We conduct searching on CIFAR10 and CIFAR100 and show architecture transferability for ImageNet training. 

\textbf{Parameter-Accuracy Trade-off.} As per Sec.\ref{sec:rank_evaluation} the rank slope indicates efficacy of layer learning. We can  then assign a threshold, $\delta$ to inform our choice of rank-slope. In choosing $\delta$, CONet considers the information gain compared to the number of additional channels, implicitly balancing accuracy gain and model size. A small $\delta$ yields larger parameter counts, and larger $\delta$ yields smaller parameter (param) counts. As in Table \ref{table_results}, DARTS$14$ searched on CIFAR100 with $\delta_2$ reduces the param-count 3.1M while using $\delta_0$ increases the param-count to 12.7M. In particular, ResNet CIFAR10 with $\delta_2$ maintains a similar accuracy, while reducing param-counts from 21.28M to 5.45M, representing a \textbf{4x reduction}. This demonstrates CONet's ability to remove redundant channel sizes without losing performance. Furthermore, DARTS$7$ achieves $6.94\%$ accuracy gain with $\delta_6$ compared to the baseline, indicating the ability of CONet to further improve the performance of the architecture through dynamic channel sizes adjustment.

\textbf{Ablation Study of Hyperparameters (HP)}: We highlight that the $\delta$ is the main HP for searching CONet. The rest of HPs fall in to three categories: (1) Convergence (i.e epochs-per-trial,  total trials, \& $\phi$), decreasing these HPs will produce an ``early'' result; (2) Limits: 
bound ranges for parameter counts; and (3) Placeholder values ($\gamma$, $\mu$) maintain CONet's robustness. Table \ref{table_ablative_study} studies how changing epochs-per-trial and $\phi$, leads to different model sizes. Changing these values only affects the search range for $\delta$ in turn controlling model size.
\begin{table}[htp]
    \setlength\tabcolsep{1pt} 
    \center
    \caption{Ablative study on Epoch-per-trial and $\phi$.}
	\label{table_ablative_study}
    \scriptsize{
    \begin{tabular}{c|c|c}
    \hlinewd{1pt}
    \text{Epoch/Trial} & Param (M) & Acc (\%) \\
    \hlinewd{1pt}
    $10$ & $4.78$ & $80.65$ \\
    $15$ & $2.17$ &	$79.01$ \\
    $20$ (Default) & $1.24$	& $79.45$\\
    \hline
    \end{tabular}
    \begin{tabular}{c|c|c}
    \hlinewd{1pt}
    \text{Factor Scale} ($\phi$) & Param (M) & Acc (\%) \\
    \hlinewd{1pt}
    $0.1$ & $1.13$ & $78.16$ \\
    $0.2$ (Default) & $1.24$ & $79.45$ \\
    $0.3$ &	$1.78$ & $79.42$ \\
    $0.4$ & $2.51$ & $79.52$ \\
    \hline
    \end{tabular}
    }
    \vspace{-4mm}
\end{table}

\begin{table*}[htp]
\setlength\tabcolsep{1pt}
\centering
    \caption{Comparison of performance of \textbf{CONet} on ResNet34, DARTS$7$, DARTS+$7$, DARTS$14$ and DARTS+$14$ searched on CIFAR10/100 with different thresholds. Note $\delta_0 = 0.0025$, $\delta_1 = 0.005$, $\delta_2 = 0.0075$, $\delta_3 = 0.01$, $\delta_4 = 0.015$, $\delta_5 = 0.02$, $\delta_6 = 0.025$. We also included performance of DARTS$7$ with compound scaling \cite{tan_effnet_2019} for $\phi_1 = \sqrt{2}$, $\phi_3 = \sqrt{2} ^ 3$, $\phi_5 = \sqrt{2} ^ 5$, $\phi_7 = \sqrt{2} ^ 7$, and $\phi_9 = \sqrt{2} ^ 9$. For compound scaling, each channel is scaled by $\phi$.}
    \label{table_results}
	\tiny{
    \begin{tabular}{c|cc|cc||cc|cc}
	\hlinewd{1.3pt}
    \multicolumn{1}{c|}{}&\multicolumn{4}{c||}{CIFAR10} &
    \multicolumn{4}{c}{CIFAR100}\\
	\hlinewd{1.3pt}
    \multirow{2}{*}{\textbf{Architecture}}&\textbf{Top-1}& \textbf{Params}&\textbf{Search Cost} &\multirow{2}{*}{\textbf{GPUs}}&\textbf{Top-1}& \textbf{Params}&\textbf{Search Cost} &\multirow{2}{*}{\textbf{GPUs}}\\
    &\textbf{(\%)}&\textbf{(M)}&\textbf{(GPU-days)}&&\textbf{(\%)}&\textbf{(M)}&\textbf{(GPU-days)}
    \\
    \hline
    \hline
    ResNet34(baseline)\cite{he2015deep}&$95.56_{\pm0.11}$&$21.28$&Manual&$-$&$77.88_{\pm0.43}$&$21.28$&Manual&$-$
    \\
    ResNet34($\delta_6$)&$95.12_{\pm0.08}${\color{red}($-0.44$)}&$2.05$&$0.3$&$1$&$74.69_{\pm0.02}$ {\color{red}($-3.19$)}&$2.56$&$0.3$&$1$
    \\
    ResNet34($\delta_5$)&$95.34_{\pm0.06}$ {\color{red}($-0.22$)}&$3.20$&$0.3$&$1$&$76.92_{\pm0.22}$ {\color{red}($-0.96$)}&$5.20$&$0.3$&$1$
    \\
    ResNet34($\delta_4$)&$95.59_{\pm0.13}$ {\color{green}($+0.03$)}&$5.47$&$0.4$&$1$&$78.07_{\pm0.15}$ {\color{green}($+0.19$)}&$9.33$&$0.4$&$1$
    \\
    ResNet34($\delta_3$)&$\bm{95.74_{\pm0.06}}$ {\color{green}($+0.23$)}&$10.55$&$0.5$&$1$&$\bm{78.43_{\pm0.40}}$ {\color{green}($+0.55$)}&$15.60$&$0.5$&$1$
    \\\hline
    DARTS$7$(baseline)\cite{liu2018darts}&$94.30_{\pm0.18}$&$0.20$&$4$&$1$&$73.81_{\pm0.31}$&$0.25$&$4$&$1$
    \\
    DARTS$7$($\delta_4$)&$-$ &$-$&$-$&$-$&$77.38_{\pm0.47}$ {\color{green}($+3.57$)}&$0.58$&$0.3$&$1$
    \\
    DARTS$7$($\delta_3$)&$95.18_{\pm0.17}$ {\color{green}($+0.88$)}&$0.63$&$0.3$&$1$&$79.45_{\pm0.33}$ {\color{green}($+5.64$)}&$1.24$&$0.3$&$1$
    \\
    DARTS$7$($\delta_2$)&$95.87_{\pm0.12}$ {\color{green}($+1.57$)}&$2.04$&$0.3$&$1$&$80.12_{\pm0.2}$ {\color{green}($+6.31$)}&$2.89$&$0.3$&$1$
    \\
    DARTS$7$($\delta_1$)&$96.25_{\pm0.13}$ {\color{green}($+1.95$)}&$4.11$&$0.4$&$1$&$80.53_{\pm0.08}$ {\color{green}($+6.72$)}&$5.04$&$0.4$&$1$
    \\
    DARTS$7$($\delta_0$)&$96.69_{\pm0.12}$ {\color{green}($+2.39$)}&$6.54$&$0.5$&$1$&$80.75_{\pm0.31}$     {\color{green}($+6.94$)}&$6.25$&$0.5$&$1$
    \\\hline
    DARTS$7$($\phi_1$)&$-$ 
    &$-$&Manual&$-$&$75.81_{\pm0.30}$ {\color{green}($+2.00$)}&$0.46$&Manual&$-$
    \\
    DARTS$7$($\phi_3$)&$-$ 
    &$-$&Manual&$-$&$78.77_{\pm0.14}$ {\color{green}($+4.96$)}&$1.67$&Manual&$-$
    \\
    DARTS$7$($\phi_5$)&$96.16_{\pm0.094}$ {\color{green}($+1.89$)}&$3.85$&Manual&$-$&$80.27_{\pm0.25}$ {\color{green}($+6.46$)}&$3.94$&Manual&$-$
    \\
    DARTS$7$($\phi_7$)&$96.32_{\pm0.16}$ {\color{green}($+2.02$)}&$5.66$&Manual&$-$&$80.50_{\pm0.25}$     {\color{green}($+6.85$)}&$5.76$&Manual&$-$
    \\
    DARTS$7$($\phi_9$)&$-$ 
    &$-$&Manual&$-$&$80.66_{\pm0.21}$     {\color{green}($+6.94$)}&$6.96$&Manual&$-$
    \\\hline
    DARTS+$7$(baseline)\cite{liang2020darts}&$94.31_{\pm0.07}$&$0.26$&$0.2$&$1$&$75.17_{\pm0.31}$&$0.28$&$0.2$&$1$
    \\
    DARTS+$7$($\delta_3)$&$95.21_{\pm0.11}$ {\color{green}($+0.9$)}&$0.81$&$0.3$&$1$&$79.87_{\pm0.27}$ {\color{green}($+4.7$)}&$2.16$&$0.3$&$1$
    \\
    DARTS+$7$($\delta_2$)&$95.84_{\pm0.15}$ {\color{green}($+1.53$)}&$1.38$&$0.3$&$1$&$80.67_{\pm0.34}$ {\color{green}($+5.5$)}&$3.54$&$0.3$&$1$
    \\
    DARTS+$7$($\delta_1$)&$96.19_{\pm0.12}$ {\color{green}($+1.88$)}&$4.2$&$0.4$&$1$&$81.01_{\pm0.13}$ {\color{green}($+5.84$)}&$5.61$&$0.4$&$1$
    \\
    DARTS+$7$($\delta_0$)&$96.46_{\pm0.11}$ {\color{green}($+2.15$)}&$7.05$&$0.5$&$1$&$81.21_{\pm0.21}$ {\color{green}($+6.04$)}&$7.44$&$0.5$&$1$
    \\\hline
    DARTS$14$(baseline)\cite{liu2018darts}&$96.53_{\pm0.12}$&$3.93$&$4$&$1$&$81.43_{\pm0.17}$&$4.00$&$4$&$1$
    \\
    DARTS$14$($\delta_3$)&$94.46_{\pm0.12}${\color{red}($-2.07$)}&$0.17$&$0.8$&$1$&$76.33_{\pm0.33}${\color{red}($-5.1$)}&$0.41$&$0.8$&$1$
    \\
    DARTS$14$($\delta_3$)&$95.54_{\pm0.1}${\color{red}($-0.99$)}&$0.46$&$0.8$&$1$&$78.4_{\pm0.31}${\color{red}($-3.03$)}&$0.91$&$0.8$&$1$
    \\
    DARTS$14$($\delta_2$)&$96.24_{\pm0.14}${\color{red}($-0.29$)}&$1.38$&$0.8$&$1$&$80.67_{\pm0.41}${\color{red}($-0.76$)}&$3.06$&$0.8$&$1$
    \\
    DARTS$14$($\delta_1$)&$96.6_{\pm0.08}${\color{green}($+0.07$)}&$4.07$&$0.9$&$1$&$82.06_{\pm0.33}${\color{green}($+0.63$)}&$8.2$&$0.9$&$1$
    \\
    DARTS$14$($\delta_0$)&$96.79_{\pm0.09}${\color{green}($+0.26$)}&$10.42$&$1$&$1$&$82.4_{\pm0.57}${\color{green}($+0.97$)}&$12.65$&$1$&$1$
    \\\hline
    
    DARTS+$14$(baseline)\cite{liang2020darts}&$96.43_{\pm0.09}$&$3.35$&$0.2$&$1$&$80.84_{\pm0.89}$&$3.40$&$0.2$&$1$
    \\
    DARTS+$14$($\delta_4$)&$94.18_{\pm0.17}${\color{red}(-2.25)}&$0.17$&$0.8$&$1$&$76.3_{\pm0.49}${\color{red}($-4.89$)}&$0.59$&$0.8$&$1$
    \\
    DARTS+$14$($\delta_3$)&$95.08_{\pm0.08}${\color{red}(-1.34)}&$0.44$&$0.8$&$1$&$78.1_{\pm0.14}${\color{red}($-2.73$)}&$1.60$&$0.8$&$1$
    \\
    DARTS+$14$($\delta_2$)&$95.78_{\pm0.09}${\color{red}(-0.64)}&$1.19$&$0.8$&$1$&$79.46_{\pm0.33}${\color{red}($-1.37$)}&$2.89$&$0.8$&$1$
    \\
    DARTS+$14$($\delta_1$)&$96.5_{\pm0.19}${\color{green}(0.07)}&$3.59$&$0.9$&$1$&$80.68_{\pm0.25}${\color{red}($-0.15$)}&$5.46$&$0.9$&$1$
    \\
    DARTS+$14$($\delta_0$)&$96.69_{\pm0.16}${\color{green}(0.26)}&$7.01$&$1$&$1$&$81.47_{\pm0.26}${\color{green}($+0.64$)}&$8.57$&$1$&$1$
    \\\hlinewd{1pt}
		\end{tabular}
    }
\end{table*}

\begin{table*}[htp]
    \setlength\tabcolsep{1pt} 
    \center
	\caption{Comparison with DARTS-related architectures on ImageNet ordered by novelty. DARTS$7$ and DART$14$ are DARTS(2nd) 7 cells and 14 cells respectively. Note $\delta_0 = 0.0025$, $\delta_1 = 0.005$, $\delta_2 = 0.0075$, $\delta_3 = 0.01$, $\delta_4 = 0.015$, $\delta_5 = 0.02$, $\delta_6 = 0.025$.}
	\label{tab:imagenet_results}
	\tiny{
		\begin{tabular}{c|ccc|ccc||ccccc}
		\hlinewd{1pt}
		&\multicolumn{3}{c|}{\textbf{ImageNet Results}}&\multicolumn{3}{c||}{\textbf{Architecture Search}}&\multicolumn{4}{c}{\textbf{ImageNet Training Setup}}\\
		\hlinewd{1pt}
		\multirow{2}{*}{\textbf{Architecture}}&\textbf{Top-1}&\textbf{Top-5}&\textbf{Params}&\textbf{Search Cost}&\textbf{Searched On}&\multirow{2}{*}{\textbf{GPUs}}&\textbf{Training}&\textbf{Batch} &\multirow{2}{*}{\textbf{Optimizer}}&\multirow{2}{*}{\textbf{LR-Schedular}}\\
		&\textbf{(\%)}&\textbf{(\%)}&\textbf{(M)}&\textbf{(GPU-days)}&\textbf{Dataset}&&\textbf{ Epochs}&\textbf{Size}\\
		\hline
		\hline
        NASNet-A\cite{zoph2018learning}&$74.0$&$91.6$&$5.3$&3-4&CIFAR10&$450$&$250$&$128$&SGD&StepLR: Gamma=0.97,Step-size=1\\
        AmoebaNet-C\cite{real2019regularized}&$75.7$&$92.4$&$6.4$&$7$&CIFAR10&$250$&$250$&$128$&SGD&StepLR: Gamma=0.97,Step-size=2\\
        DARTS-7\cite{liu2018darts}&$52.9$&$76.9$&$0.5$&$4$&CIFAR10&$1$&200&128&SGD&StepLR: Gamma=0.5,Step-size=25\\
        DARTS-14\cite{liu2018darts}&$73.3$&$91.3$&$4.7$&$4$&CIFAR10&$1$&250&128&SGD&StepLR: Gamma=0.97,Step-size=1\\
        GHN\cite{zhang2018graph}&$73.0$&$91.3$&$6.1$&$0.84$&CIFAR10&$1$&$250$&$128$&SGD&StepLR: Gamma=0.97,Step-size=1\\
        ProxylessNAS\cite{cai2018proxylessnas}&$74.6$&$92.2$&$5.8$&$8.3$&ImageNet&$-$&$300$&256&Adam &$-$\\
        SNAS\cite{xie2018snas}&$72.7$&$90.8$&$4.3$&$1.5$&CIFAR10&$1$&$250$&$128$&SGD&StepLR: Gamma=0.97,Step-size=1 \\
        BayesNAS\cite{zhou2019bayesnas}&$73.5$&$91.1$&$3.9$&$0.2$&CIFAR10&$1$&$250$&$128$&SGD&StepLR: Gamma=0.97,Step-size=1\\
        P-DARTS\cite{chen2019progressive}&$75.6$&$92.6$&$4.9$&$0.3$&CIFAR10&8&250&1024&SGD&OneCycleLR: Linear Annealing\\
        GDAS\cite{dong2019searching}&$72.5$&$90.9$&$4.4$&$0.17$&CIFAR10&1&250&128&SGD&StepLR: Gamma=0.97,Step-size=1\\
        PC-DARTS\cite{xu2019pc}&$75.8$&$92.7$&$5.3$&$3.83$&ImageNet&8&250&1024&SGD&OneCycleLR: Linear Annealing\\
        DARTS+\cite{liang2020darts}&$76.3$&$92.8$&$5.1$&$0.2$&CIFAR100&1&800&2048&SGD&OneCycleLR: Cosine Annealing\\
        NASP\cite{yao2020efficient}&$73.7$&$91.4$&$9.5$&$0.1$&CIFAR10&$1$&$250$&$128$&SGD&StepLR: Gamma=0.97,Step-size=1\\
        SGAS\cite{li2020sgas}&$75.9$&$92.7$&$5.4$&$0.25$&CIFAR10&1&250&1024&SGD&OneCycleLR: Linear Annealing\\
        MiLeNAS\cite{he2020milenas}&$75.3$&$92.4$&$5.3$&$0.3$&CIFAR10&$1$&250&128&SGD&StepLR: Gamma=0.97,Step-size=1\\
        SDARTS\cite{chen2020stabilizing}&$75.3$&$92.2$&$3.3$&$1.3$&CIFAR10&1&250&1024&SGD&OneCycleLR: Linear Annealing\\
        FairDARTS\cite{chu2020fair}&$75.6$&$92.6$&$4.3$&$3$&ImageNet&1&250&1024&SGD&OneCycleLR: Linear Annealing\\
        \hline
        DARTS$7$($\delta_{3}$)&$67.9$&$88.0$&$1.8$&$0.3$&CIFAR100&1&200&128&SGD&StepLR: Gamma=0.5,Step-size=25\\
        DARTS$7$($\delta_{2}$)&$72.0$&$90.6$&$3.4$&$0.4$&CIFAR100&1&200&128&SGD&StepLR: Gamma=0.5,Step-size=25\\
        DARTS$7$($\delta_{1}$)&$74.1$&$91.7$&$5.6$&$0.5$&CIFAR100&1&200&128&SGD&StepLR: Gamma=0.5,Step-size=25\\
        DARTS-14($\delta_{2}$)&$67.2$&$87.6$&$1.8$&$0.8$&CIFAR10&1&200&128&SGD&StepLR: Gamma=0.97,Step-size=1\\
        DARTS-14($\delta_{1}$)&$74.0$&$91.8$&$4.8$&$0.8$&CIFAR10&1&200&128&SGD&StepLR: Gamma=0.97,Step-size=1\\
        \textbf{DARTS$14$($\delta_{1}$)}&$\textbf{76.6}$&$\textbf{93.2}$&$9.0$&$0.9$&CIFAR100&1&200&128&SGD&StepLR: Gamma=0.5,Step-size=25\\
        
        \hlinewd{1pt}
		\end{tabular}
    }
\end{table*}

\textbf{Comparison to Compound Scaling:} EfficientNet's compound scaling \cite{tan_effnet_2019} heuristically increases the channel sizes relative to the network depth.  This poses challenges where it can unnecessarily increase the number of channels per layer without consideration of (1) the information that has been learned and (2) the amount of redundant channel sizes. The CONet Alg. \ref{alg:coalg} is able to inspect weights to observe how well the model is learning and dynamically increase \textit{or} decrease the number of channels (if necessary), allowing the network to evolve itself to the optimal number of channels. As shown in Fig.\ref{fig:size_vs_acc}, CONet remains superior compared to compound scaling on CIFAR/DARTS$7$ related experiments.


\textbf{Architecture Transferability:} The DARTS$7$ and DARTS$14$ architectures searched on CIFAR10/100 are evaluated on ImageNet to test architecture transferability, seen in Table \ref{tab:imagenet_results}. We compare the performance of our searched models to their baseline as well as other SOTA methods with a similar search space to DARTS. Comparing against the baseline, we see a $21.2\%$ increase in performance for DARTS-7 $\delta_1$ and $2.9\%$ increase in performance for DARTS-14 $\delta_1$. This demonstrates that our larger model is also capable of achieving a higher performance than our smaller models on ImageNet, reflecting the performance gain as shown on CIFAR10/100. 

\textbf{Generalization:} In addition to DAG-based relaxed search spaces like DARTS and traditional CNNs like ResNet, the methodology of CONet can be applied to any type of CNN. We have demonstrated the ability of CONet to balance performance and parameters and shown the clear merits of our method to dynamically \textit{scale-up} or \textit{scale-down} an architecture as compared to heuristic channel size assignments and EfficientNet compound scaling. This method can be applied co-currently to other architecture search that trains a network during search. The methodology we propose provides an indicator to probe the useful information learned by a CNN layer, which can then be used to facilitate channel size selection during architecture search. \\


\vspace{-0.2cm}
\textbf{Rank correlation to test-accuracy}: The Rank metric closely follows test accuracy while \textit{only being exposed to train data}. In Fig \ref{fig:Concept_Figure}(b) the layer rank follows test accuracy for 250 epochs. Additionally, our CONet Alg. \ref{alg:coalg} operates in the growth phase of training, seen in Fig \ref{fig:Concept_Figure}(b) bottom right, where the rank measure is directly proportional to the test accuracy. The NATS benchmark \cite{dong2021nats} is composed of 15, 625 neural architectures, in Fig. \ref{fig:Concept_Figure}(c) we inspect the rank of each at the 90th epoch \& show it is highly correlated to test-accuracy with a Spearman correlation coefficient of 0.913. Therefore, rank as a novel metric allows channel optimizations from only train data.

\section{Conclusion}\label{sec:conclusion}
In this work, we studied an important problem in NAS: channel size optimization given a specific model and task. To this end, we proposed CONet; a dynamic scaling algorithm that uses the \textit{Rank} measurements from low-rank factorized tensor weights to efficiently scale channel sizes layers to achieve higher performance. We demonstrated that layer \textit{Rank} takes a few training cycles to stabilize and we can use the average \textit{Rank} of multiple layers to evaluate their performance as a whole. By representing the model as \textit{conv} and \textit{non-conv} connectors, we can automatically adjust the channel sizes of any given CNN with any dataset. Furthermore, we have shown that the adjustments CONet makes is \textit{better} than uniformly scaling channel sizes with compound scaling method. In the future, we plan on extending CONet to kernel size optimization and to develop deeper integration with other popular NAS techniques so that they may run concurrently during architecture search.

{\small
\bibliographystyle{ieee_fullname}
\bibliography{egbib}
}

\clearpage
\appendix
\tableofcontents
\clearpage

\section{Appendix A: Experiment setup Details}\label{appendix_A}
\subsection{General Experiment Setup}
For our experiments, we used the CIFAR10, CIFAR100 and ImageNet (ILSVRC2012) \cite{krizhevsky2009learning, deng2009imagenet} datasets. The CIFAR10 and CIFAR100 datasets have 50k training images with 10 and 100 classes with 5000 and 500 images per class respectively. The ImageNet dataset has 1.2M training images with 1000 classes.

We search and evaluate models on CIFAR datasets using one Nvidia V100. For architecture transfer to ImageNet, we evaluate smaller models using one Nvidia RTX 2080Ti and larger models using two Nvidia RTX 2080Tis to accommodate model size. 

\subsection{Experiment setup for channel search}
Before the channel search, we set the channel sizes of the model to 32 for each layer. For the DARTS$7$ and DARTS+$7$ experiments on CIFAR10, we used the original channel sizes for our initial model. In order to facilitate optimal rank growth, the initial learning rate $\eta$ is a valuable hyper-parameter, therefore for DARTS and DARTS+, we chose $0.1$, and for ResNet34, we chose $0.03$. 

Within the search phase, we used the Adas\footnote{https://github.com/mahdihosseini/AdaS} scheduler \cite{anonymous2020adas} with $\beta = 0.8$ for channel search. Adas is an adaptive learning rate scheduler that also uses local indicators to individually adjust the learning rate of each convolutional layer. $\beta$ is the learning momentum factor used by Adas. We found that setting $\beta$ to 0.8 led to quick stabilization of rank and mapping condition while maintaining the underlying structure for each convolutional layer.

\subsection{Experiment setup: CIFAR Evaluation}
We use the stochastic gradient descent optimizer (SGD) with momentum of $0.9$ and a step decaying learning rate scheduler (StepLR) with a step size of $25$ epochs and step decay of $0.5$. For all evaluations, we set the initial learning rate, $\eta$, to $0.1$. For ResNet34 model evaluation, we used a weight decay of $0.0005$. For DARTS/DARTS+ model evaluation, we used a weight decay of $0.0003$ and all of the additional hyperparameters outlined in \cite{liu2019darts} (e.g. auxiliary head/weight, cutout). Every model is evaluated for $250$ epochs for at least $3$ runs and the averaged results are reported. 
 
 \subsection{Experiment setup: ImageNet Evaluation}
For DARTS-14 searched on CIFAR100 and DARTS-7 searched on CIFAR10/100, we used the same settings as the evaluation of DARTS/DART+ models on CIFAR except the weight decay is set to $0.00003$. For DARTS-14 searched on CIFAR10 with $\delta_1$, $\delta_2$, we additionally use step size of 1 and a decay rate of 0.97 for the StepLR scheduler. Each model is evaluated for 200 epochs for 1 run.


\section{Appendix B: Further Experiment Result}\label{appendix_B}

\begin{figure*}[htp!]
\vspace*{-0.6cm}
\centering
    \includegraphics[width=\textwidth]{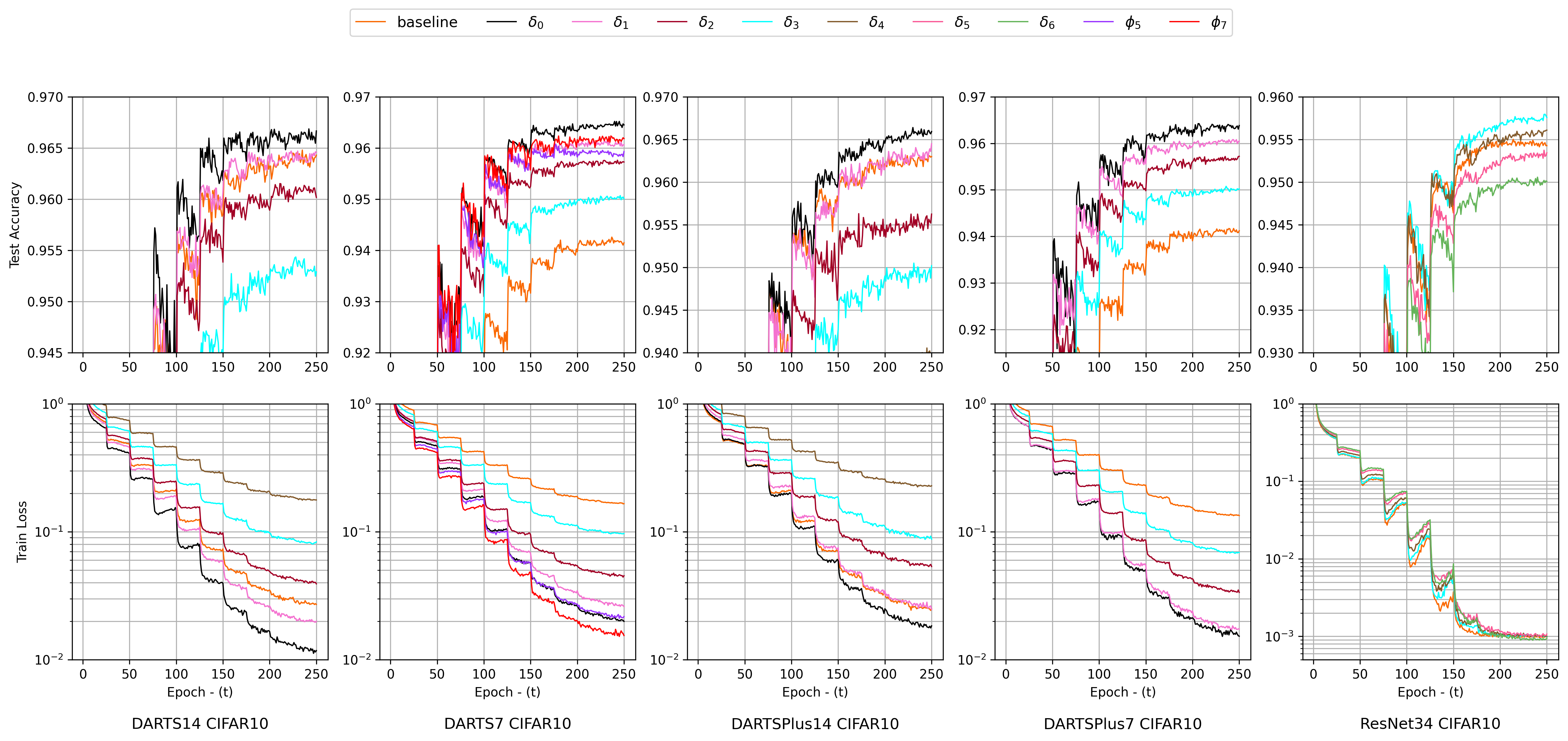}
\caption{Full training results on CIFAR10 for DARTS-7, DARTS-14, DARTS+-7, DARTS+-14 and ResNet searched on CIFAR10 with different delta thresholds: $\delta_0 = 0.0025$, $\delta_1 = 0.005$, $\delta_2 = 0.0075$, $\delta_3 = 0.01$, $\delta_4 = 0.015$, $\delta_5 = 0.02$, $\delta_6 = 0.025$. Compound Scaling for DARTS7 with different Phi values: $\phi_5 = \sqrt{2} ^ 5$, $\phi_7 = \sqrt{2} ^ 7$.}
\label{fig:cifar10_full_results}
\end{figure*}

\begin{figure*}[htp!]
\vspace*{-0.6cm}
\centering
    \includegraphics[width=\textwidth]{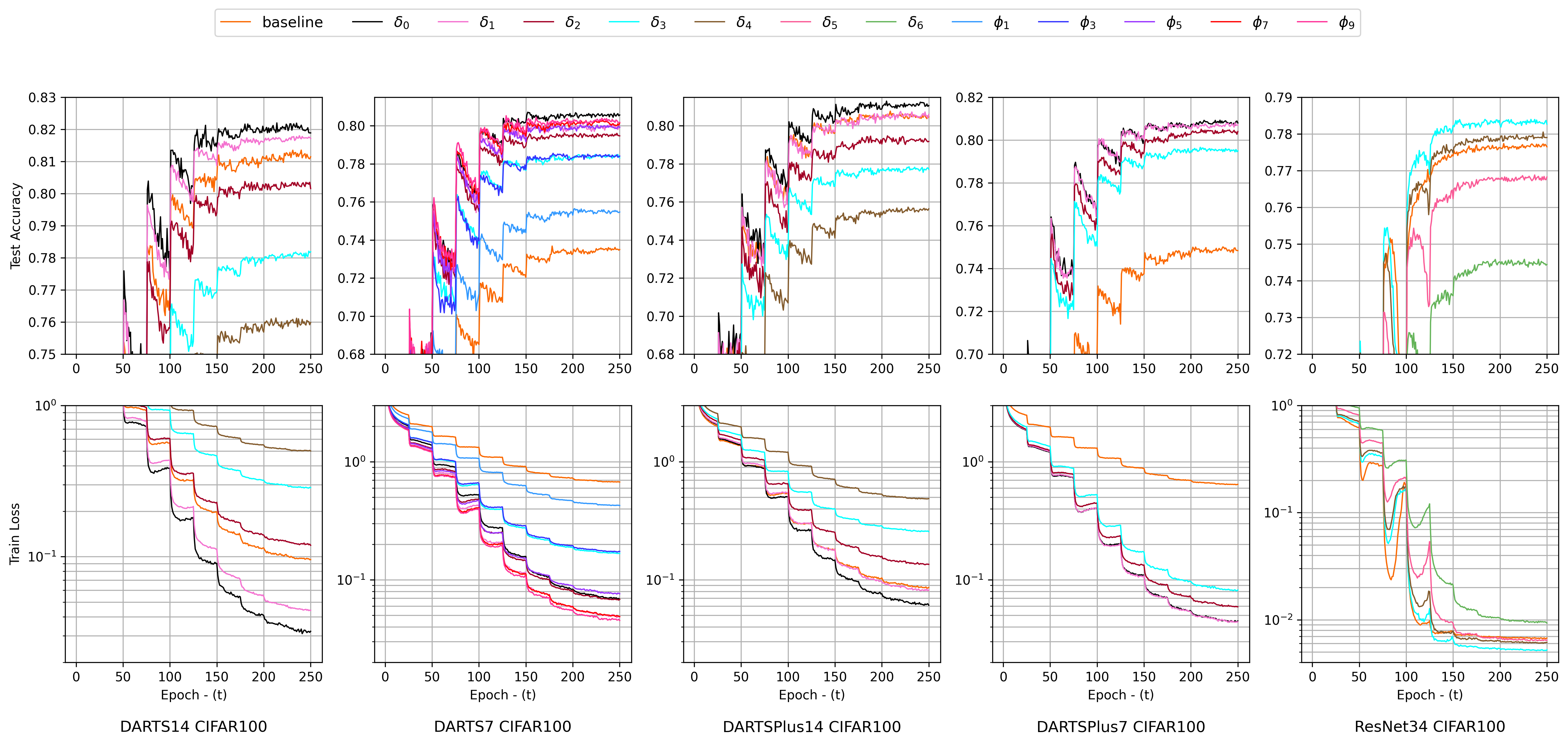}
\caption{Full training results on CIFAR100 for DARTS-7, DARTS-14, DARTS+-7, DARTS+-14 and ResNet searched on CIFAR100 with different delta thresholds: $\delta_0 = 0.0025$, $\delta_1 = 0.005$, $\delta_2 = 0.0075$, $\delta_3 = 0.01$, $\delta_4 = 0.015$, $\delta_5 = 0.02$, $\delta_6 = 0.025$. Compound Scaling for DARTS7 with different Phi values: $\phi_1 = \sqrt{2}$, $\phi_3 = \sqrt{2} ^ 3$, $\phi_5 = \sqrt{2} ^ 5$, $\phi_7 = \sqrt{2} ^ 7$, and $\phi_9 = \sqrt{2} ^ 9$.}
\label{fig:cifar100_full_results}
\end{figure*}

\begin{figure*}[htp!]
\vspace*{-0.6cm}
\centering
    \includegraphics[width=\textwidth]{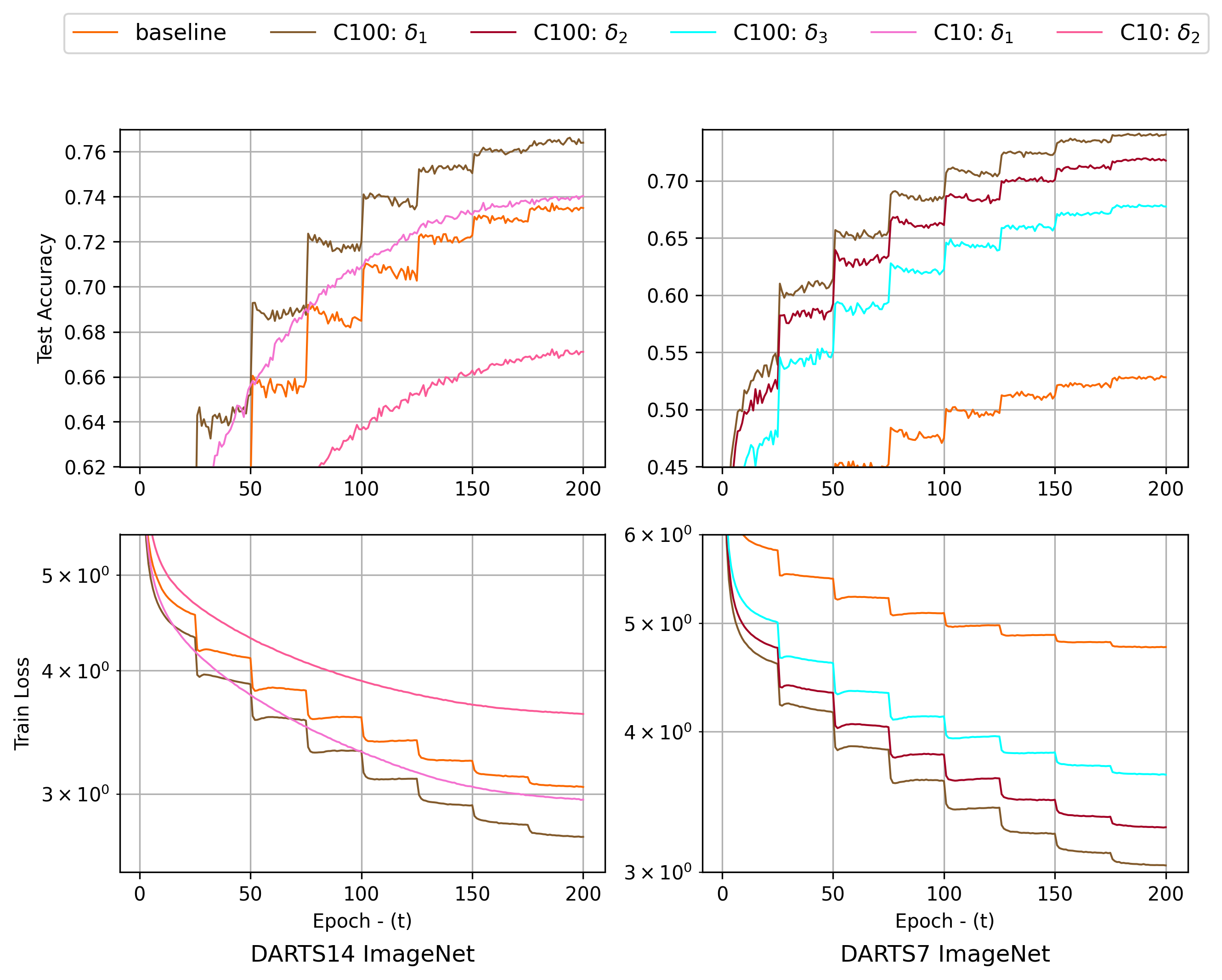}
\caption{Full training results on ImageNet for DARTS-7 and DARTS-14 searched on CIFAR10/100 with different delta thresholds: $\delta_1 = 0.005$, $\delta_2 = 0.0075$, $\delta_3 = 0.01$, $\delta_4 = 0.015$.}
\label{fig:imagenet_full_results}
\end{figure*}

We have included the training curves for all model evaluations reported in {\color{red}subsection 5.1}. For each evaluation, we plot the training loss and test accuracy per epoch. We categorize the results by dataset. For CIFAR experiments, we report the average train loss and test accuracy across multiple runs. Please see \autoref{fig:cifar10_full_results} for CIFAR10 plots, \autoref{fig:cifar100_full_results} for CIFAR100 plots, and \autoref{fig:imagenet_full_results} for ImageNet plots.

Please note that for the ImageNet experiments with the DARTS14 models searched on CIFAR10, we used a different step size and decay rate for the StepLR scheduler in contrast to our other experiments. Following the training procedure outline in \cite{liu2019darts}, we used a step size of 1 and decay rate of 0.97.

\end{document}